\documentclass[11pt,letterpaper]{arxiv}
\pdfoutput=1
\usepackage[numbers,sort&compress]{natbib}
\usepackage{parskip}
\usepackage{arxiv}

\title{GOTT: Object-centric Dexterous Manipulation with a Reusable Cross-Embodiment Primitive
}

\author[1,2]{Yulin Liu}
\author[1,3]{Lai Wei}
\author[1,4]{Yen-Jen Wang}
\author[1]{Akash Sharma}
\author[1,4]{Pieter Abbeel}
\author[2]{Henrik I. Christensen}
\author[1,3]{Haozhi Qi}

\affil[1]{Amazon FAR}
\affil[2]{UC San Diego}
\affil[3]{University of Chicago}
\affil[4]{UC Berkeley}

\projectwebsite{https://dex-gott.github.io/}{\texttt{https://dex-gott.github.io}}

\renewcommand{\preabstractcontent}{%
\begin{center}
    \includegraphics[width=\linewidth]{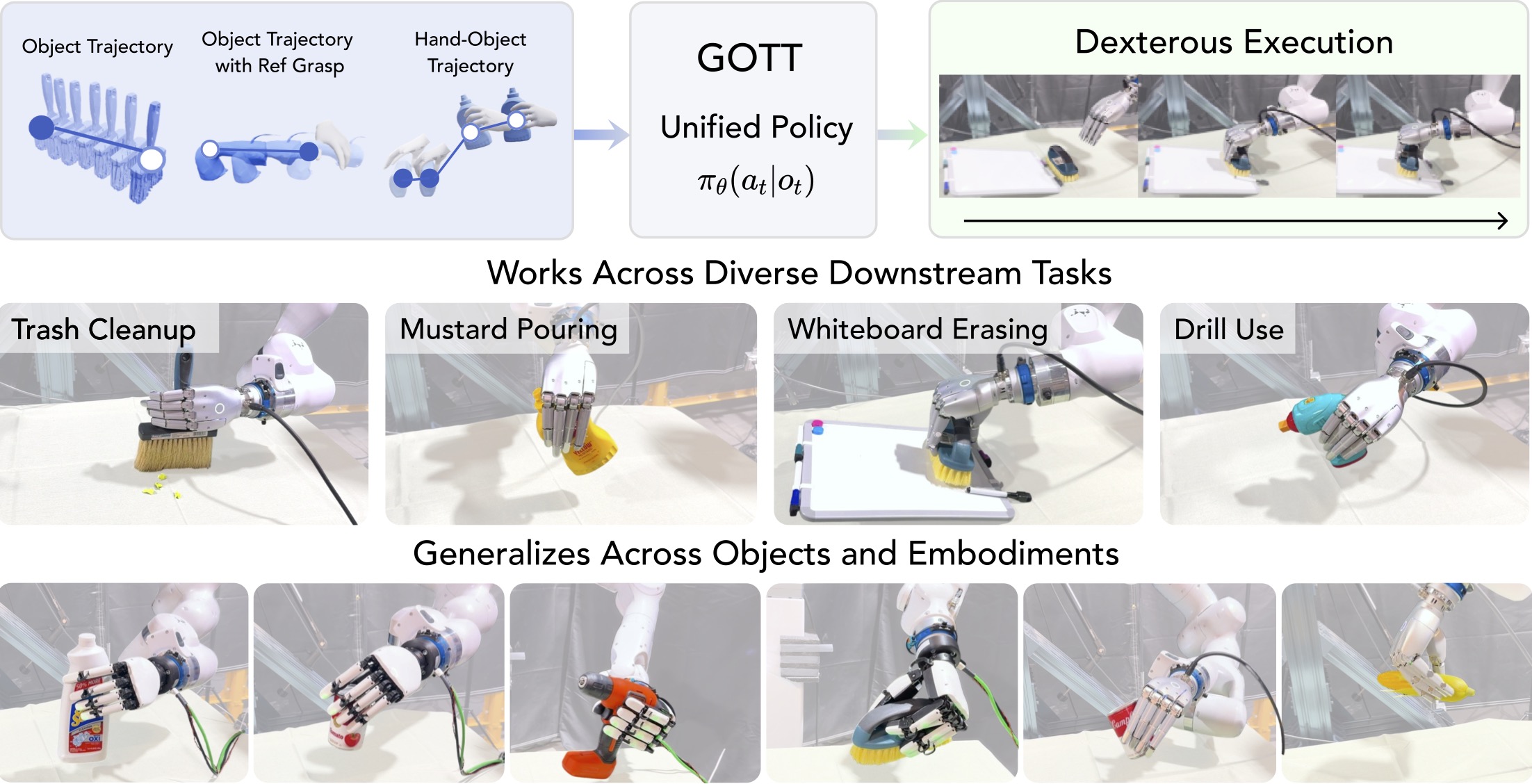}
    \captionof{figure}{\ours converts object trajectories into dexterous robot execution, with or without additional guidance from reference grasps or hand-object trajectories. The same contact-acquisition policy supports object manipulation across hand embodiments and diverse downstream tasks.}
    \label{fig:teaser}
\end{center}
}

\begin{document}

\begin{abstract}
Foundation models and large-scale human data provide rich sources of manipulation intent, but translating this intent into multi-fingered robot behavior remains difficult. Dexterous hands still lack a \textbf{reusable} low-level primitive that reliably establishes contact across tasks and embodiments. We propose \ours, a \emph{reach-acquire-move} framework built around a single cross-embodiment contact-acquisition primitive. Given a robot-agnostic object trajectory and a reach specification, \ours first brings the hand near a task-relevant contact region. The shared closed-loop primitive then establishes stable contact from this approximate initialization, and a pose-conditioned controller tracks the desired object motion. Reach specifications may come from future-aware planning, external models, or human demonstrations, while the primitive and tracking backend remain unchanged. Simulation and real-world experiments show that \ours is able to establish robust contact across diverse objects, arm-hand platforms, and seen and unseen hand morphologies. It also consistently improves end-to-end task success over open-loop grasp execution.

\end{abstract}

\maketitle

\secv
\section{Introduction}
\secv
  
Foundation models~\cite{deepmind2026gemini31pro,openai2026gpt55} and large-scale human data~\cite{yang2022oakink,banerjee2024introducing} provide rich sources of manipulation intent. For parallel-jaw grippers, such intent can be translated into robot actions through mature low-level primitives such as grasping, placing, and trajectory following~\cite{huang2023voxposer,huang2024rekep,liang2023code}. However, the same approach does not directly extend to dexterous hands, which lack a comparable set of general-purpose low-level skills. These skills must establish complex contact configurations, coordinate high-dimensional actions, and adapt to different hand morphologies.

Existing dexterous manipulation methods usually address these challenges through task-specific sim-to-real learning~\cite{andrychowicz2020learning} or hand-motion retargeting~\cite{li2025maniptrans}. However, these approaches are not well suited to converting high-level intent into robot behavior. Task-specific policies~\cite{qi2022hand,qi2023rotateit,chen2023visual,nvidia2022dextreme} require retraining for each new behavior, making them difficult to scale. Retargeting recovers kinematic targets from demonstrations~\cite{mandi2025dexmachina}, often without incorporating perceptual feedback~\cite{pan2025spider}. This lack of feedback makes contact-rich tasks difficult to execute in the real world. This raises a key question: \textit{what is the right interface for translating high-level manipulation intent into low-level dexterous execution?}

Object-centric motion representations provide a promising answer to this question. Prior work has shown that many rigid-object manipulation tasks can be described by the desired object motion~\cite{kerr2024robot,patel2025robotic,heppert2024ditto}, abstracting away embodiment-specific robot motion. However, an object trajectory specifies \emph{how the object should move}, but not \emph{where and how the hand should establish contact}. This distinction is critical for functional manipulation: a geometrically stable grasp may still be unsuitable for pouring, sweeping, or tool use.
Prior methods address this gap by supplying a pre-grasp or hand reference~\cite{chen2024object,liu2025dextrack}, or by learning contact acquisition jointly with trajectory execution~\cite{kedia2026simtoolreal,kuang2026dex4d}. While effective, these designs do not expose contact acquisition as a reusable skill across high-level sources and robot embodiments.

We propose \textbf{General Object Trajectory Tracking} (\ours), a framework for executing diverse object motions across tasks and robot embodiments using a shared contact-acquisition primitive. 
We represent high-level manipulation intent using a desired object trajectory and a \emph{reach specification}, which specifies how to bring the hand near a task-relevant contact region. These components may come from different sources. For example, the object trajectory may be extracted from a video, while the reach specification may be generated through affordance-aware planning or obtained by retargeting a human trajectory.

\ours implements this interface through a \emph{reach, acquire, and move} decomposition. During \textit{reach}, the robot follows the reach specification toward the selected contact region. During \textit{acquire}, a learned closed-loop contact-acquisition primitive establishes stable contact from this approximate initialization. During \textit{move}, an object-pose-conditioned controller tracks the desired object trajectory. The primitive uses only local hand-object geometry and proprioceptive history, while a canonical per-link representation allows the same policy to be shared across hands. Reach specifications can come from VLM-guided, future-aware planning, external models, or demonstrations, while the contact-acquisition primitive remains unchanged.

We evaluate \ours in simulation and the real world. First, we test whether the contact-acquisition primitive can robustly grasp diverse objects across different hands and transfer zero-shot to unseen hand morphologies. We then evaluate the full pipeline on object-trajectory tracking tasks, including trash cleanup, pouring, whiteboard erasing, and tool use. We compare against end-to-end trajectory learning, showing that our decomposition enables robustness to workspace shifts and reuse across robot arms. We also show that the same primitive supports high-level specifications from generated videos, keyframes, human demonstrations, and hand-object interaction datasets. Finally, we conduct ablations on the primitive's observation and reward design and the system's planning modules.

\section{Related Work}

\paragraphc{High-level task interfaces.} Foundation models and human demonstrations can specify robot manipulation tasks through language, visual constraints~\cite{huang2024rekep}, code~\cite{liang2023code}, or generated videos~\cite{patel2025robotic}. ManipGen~\cite{dalal2024local} combines reusable local policies with foundation models and motion planning. Another line of work learns from human videos by extracting intermediate motion representations, such as point trajectories, object flow, or object-centric motion plans~\cite{bharadhwaj2024track2act,wen2023any,xu2024flow,zhu2026vision,heppert2024ditto,li2024okami}. SPOT~\cite{hsu2024spot} learns object-pose trajectory plans from human demonstrations and uses them to condition a robot policy. VLM-Scaffold~\cite{de2026scaffolding} adapts a similar pipeline to dexterous hands but relies on task-specific residual RL, making it challenging to scale across tasks and embodiments. LUCID~\cite{gupta2026lucid} learns embodiment-agnostic intent from unstructured human videos and connects it to embodiment-specific sensorimotor policies through object flow and a rough palm pose. In contrast, \ours represents high-level intent using an object trajectory and a flexible reach specification, and connects these inputs to robot execution through a reusable dexterous contact-acquisition primitive.

\paragraphc{Trajectory-based dexterous manipulation.}
D-Grasp~\cite{christen2022dgrasp} combines reference-conditioned grasp control with global motion synthesis to move objects to target poses. Methods that start from a reference trajectory typically follow one of two strategies. The first performs trajectory-specific transfer, converting a human or object trajectory into robot motion through kinematic optimization, sampling~\cite{pan2025spider}, or reinforcement learning~\cite{li2025maniptrans,mandi2025dexmachina}. Human2Sim2Robot~\cite{lum2025crossing} uses a single object trajectory and a pre-grasp to learn a task-specific robot policy. ManipTrans~\cite{li2025maniptrans} and REGRIND~\cite{feng2026regrind} transfer hand-object interaction trajectories through residual learning, while SPIDER~\cite{pan2025spider} performs fast physics-informed retargeting. These transfer methods can reproduce a given demonstration but are often tied to a specific trajectory or embodiment.

The second strategy trains policies conditioned on multiple reference trajectories. ObjDex~\cite{chen2024object} and DexTrack~\cite{liu2025dextrack} condition policies on object or hand trajectories. AdaDexTrack~\cite{adalibieke2026adadextrack} adds online modulation to improve tracking of noisy references. OmniGrasp~\cite{luo2024omnigrasp} learns to grasp diverse objects and follow object trajectories with a simulated humanoid. SimToolReal~\cite{kedia2026simtoolreal} and Dex4D~\cite{kuang2026dex4d} learn general object-centric policies conditioned on target object poses or point tracks, jointly handling contact acquisition and motion execution. Grasp to Act~\cite{gupta2026grasp2act} combines human-guided grasp optimization with residual RL for online finger adaptation.
In contrast, \ours factors execution into \emph{reach, acquire, and move}: high-level inputs specify the desired object motion and a task-relevant reach, while a trajectory-agnostic primitive establishes contact before trajectory tracking begins. This decomposition enables the same contact-acquisition primitive to be reused across trajectories, objects, and robot embodiments.

\paragraphc{Dexterous grasping.} Existing learning-based methods develop generalizable tabletop grasping policies~\cite{xu2023unidexgrasp,wan2023unidexgrasppp,lum2024dextrah,zhang2025robustdexgrasp}. Grasp synthesis and evaluation methods instead generate or refine grasp targets~\cite{chen2025bodex,wei2024dro,wang2022dexgraspnet,chen2024springgrasp,lum2024get,fang2025anydexgrasp}. GraspXL~\cite{zhang2024graspxl} and ConGraspXL~\cite{zhang2026congraspxl} condition grasping motions on approach, contact-region, or pose constraints. Functional grasping methods use affordances, human grasp references, or language to select task-relevant grasps~\cite{agarwal2023functional,huang2025fungrasp,wei2025afforddexgrasp}. Cross-embodiment methods further share grasping policies or representations across hands: CrossDex~\cite{yuan2025cross} uses a unified observation space and retargeted eigengrasp actions, while OHRA~\cite{wei2026one} uses canonical hand representations and UniMorphGrasp~\cite{wu2026unimorphgrasp} conditions grasp generation on morphology. GET-Zero~\cite{patel2025getzero} studies zero-shot transfer to unseen hand structures for in-hand rotation. These works advance grasping and cross-hand transfer, while primarily focusing on grasp acquisition itself. In contrast, GOTT treats contact acquisition as a reusable intermediate primitive that refines an approximate, task-relevant reach into stable contact before downstream object-trajectory tracking. This enables the same primitive to be reused across objects, tasks, trajectories, and hand embodiments.

\section{Dexterous Manipulation via General Object Trajectory Tracking}

We propose \ours, a \emph{reach-acquire-move} framework (Section~\ref{sec:decomposition}) for executing rigid-object motions with dexterous hands. Its high-level interface consists of a desired object trajectory $\tau_o$ and a reach specification $\rho$ that brings the hand near a task-relevant contact region. At its core, a reusable cross-embodiment contact-acquisition primitive grounds this interface by turning an approximate reach into stable contact (Section~\ref{sec:rl}). In the object-centric setting, a future-aware planner constructs $\rho$ from the object mesh, desired trajectory, and task description (Section~\ref{sec:planning}). Alternatively, external models or human demonstrations can provide $\rho$ (Section~\ref{sec:inference}), while the primitive and tracking backend remain unchanged (Section~\ref{sec:execution}).

\subsection{Reach-Acquire-Move Decomposition}
\label{sec:decomposition}

Given a desired object trajectory $\tau_o=\{T_{O,t}\}_{t=0}^{T}$ where $T_{O,t}$ denotes the object pose at timestep $t$, and a reach specification $\rho=(\xi_{\mathrm{app}},g_{\mathrm{init}})$, \ours decomposes dexterous execution into three stages. During \emph{reach}, the robot follows the approach trajectory $\xi_{\mathrm{app}}$ to an approximate initial configuration $g_{\mathrm{init}}$. During \emph{acquire}, the contact-acquisition primitive uses closed-loop feedback to establish stable contact. During \emph{move}, the pose-conditioned tracker moves the object along $\tau_o$.

This decomposition separates high-level reasoning about functional contact selection, collision-free approach, and long-horizon reachability from local contact-rich control. Consequently, the high-level source needs to provide only a feasible approach to an approximate initial configuration, rather than an exactly executable final grasp.

\subsection{General-Purpose Contact Acquisition Primitive}
\label{sec:rl}

Establishing stable contact from an initial hand configuration near the object remains challenging.
State-of-the-art grasp-generation methods either execute grasps open loop~\cite{chen2025bodex,wei2024dro}, do not account for the functional suitability of the selected initial grasp~\cite{lum2024dextrah,zhang2025robustdexgrasp}, or do not generalize across embodiments. We therefore train a local, closed-loop contact-acquisition policy using reinforcement learning in simulation and transfer it to the real world. The policy uses local hand-object geometry and proprioceptive history, while a canonical representation enables sharing across hands. This design allows the same primitive to be reused across tasks, high-level sources, objects, and hand embodiments. We first describe the single-hand control policy and then its extension to cross-embodiment control.

\begin{figure}[t]
    \centering
    \includegraphics[width=\columnwidth]{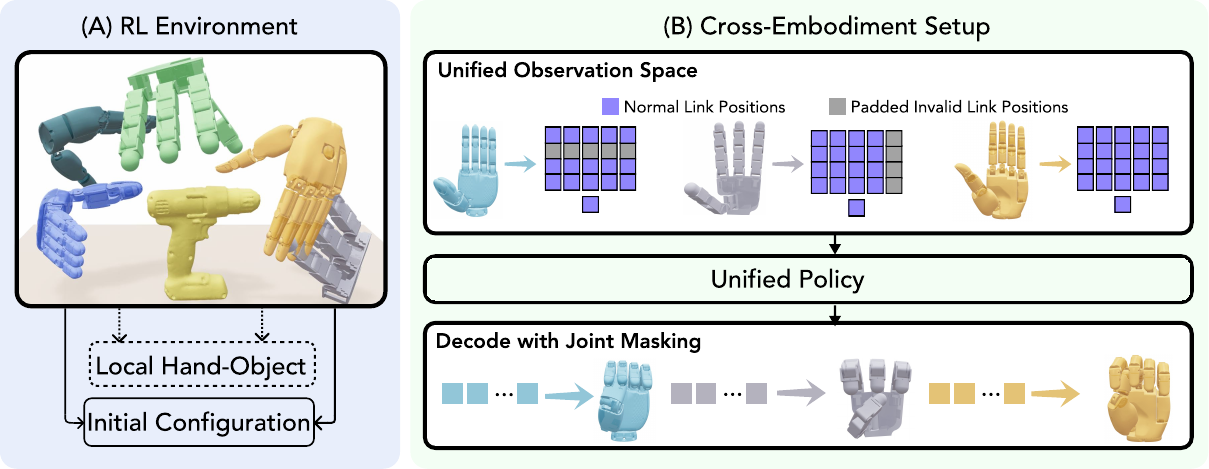}
    \caption{\small\textbf{Contact-acquisition training and cross-embodiment policy.} (A) The RL policy starts from randomized initial configurations and learns to establish stable contact using local hand-object observations. (B) Padding and binary masks map different hands into a shared observation and action space, allowing one policy to operate across hand embodiments.}
    \vspace{-1.2em}
    \label{fig:rl_and_crossdex}
\end{figure}

\paragraphc{Environment setup.} We select 16 YCB objects~\cite{calli2017yale} as our training set and collect several stable resting poses for each object. For each object--pose pair, we use BODex to generate 4,000 initial hand configurations of the type described in Section~\ref{sec:planning} and shown in Figure~\ref{fig:rl_and_crossdex}A. Each episode starts from one of these initial hand configurations. The policy must establish stable contact and lift the object.

\paragraphc{Observations and actions.} The observation contains three components: (1) the link-distance representation from~\cite{zhang2025robustdexgrasp}, in which each hand link is represented by its displacement to the nearest point in the object point cloud; (2) hand-link positions; and (3) joint positions and previous joint targets. Components (2) and (3) are stacked over the five most recent timesteps. The policy outputs a 6D wrist-delta command and a vector of hand-joint deltas: ${a}_t = \pi_\theta(o_t)$.

\paragraphc{Rewards.}
Our reward function is defined as:
$
r =
w_d r_{\mathrm{dist}}
+ w_c r_{\mathrm{contact}}
+ w_l r_{\mathrm{lift}}
+ w_o r_{\mathrm{obj}}.
$
The four terms encourage link-object proximity, opposing contact between the thumb and another finger, lifting after contact, and object stability, respectively. Additional definitions and weights are provided in the appendix.

\paragraphc{Domain randomization.}
During training, we randomize object scale, friction, and mass; robot-link friction; and actuator stiffness and damping. We also apply random force and torque perturbations once the object is held to improve grasp robustness.

\paragraphc{Contact prediction.}
During deployment, we use a proprioception-based contact predictor to determine when to switch from acquisition to tracking. Because proprioceptive and action histories can reveal contact events~\cite{qi2022hand}, we train a separate network after the policy converges:
$\hat{c}_t = f_\phi(h^{\mathrm{prop}}_{t-k+1:t}) \in [0,1]^5$,
where $h^{\mathrm{prop}}_{t-k+1:t}$ contains recent joint positions, joint targets, and hand-link positions, and $\hat{c}_t$ contains the per-finger contact probabilities. We train the network in simulation using ground-truth per-finger contact labels. Acquisition terminates when the predicted contact probabilities for the thumb and at least one other finger exceed $\tau_c$ for $k$ consecutive timesteps.

\paragraphc{Unified cross-embodiment representation.} To train one controller across multiple multi-fingered hands, we map hand-specific observations and actions into a canonical representation, as shown in Figure~\ref{fig:rl_and_crossdex}B. For link-level observations, we map each hand onto a 21-keypoint topology consisting of one palm keypoint and four keypoints per finger. Missing keypoints are filled using the nearest available links, and a binary link mask distinguishes valid entries from padded ones. Joint positions and hand actions are similarly mapped to a fixed 26-slot canonical joint order. Missing joint slots are zero-padded, and a binary joint mask identifies the active slots. The policy outputs a 6D end-effector delta and a canonical joint-delta vector. Before execution, padded joints are masked out, and the remaining entries are remapped to the robot's native joint indices. This representation allows one policy to operate across hands with different joint counts and kinematic structures without hand-specific encoder or decoder networks.

\subsection{Reach Specifications from Future-Aware Planning}
\label{sec:planning}

\begin{wrapfigure}{r}{0.6\textwidth}
    \centering
    \includegraphics[width=\linewidth]{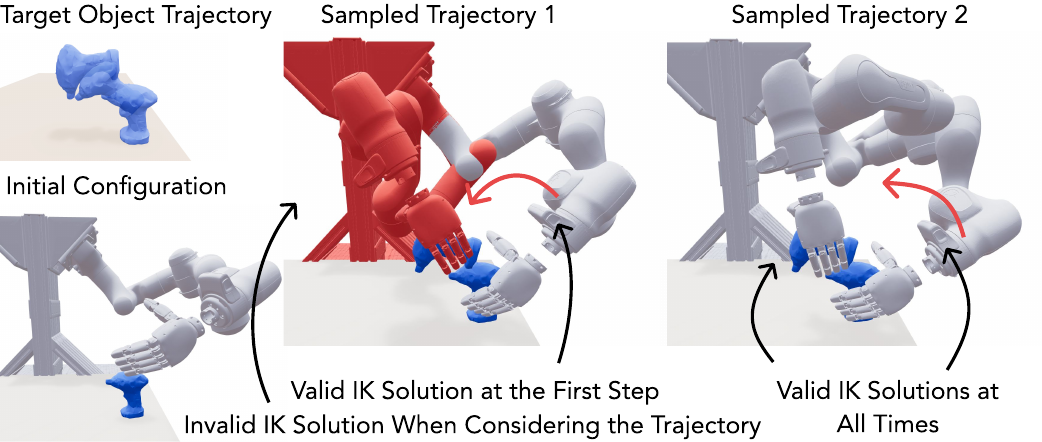}
    \caption{\small\textbf{Future-aware planning.}
    We roll out the hand motion induced by each candidate initial configuration over the full object trajectory and solve IK along this motion. Red segments indicate joint jumps from discontinuous IK solutions. We reject candidates that are initially reachable but become infeasible later and select the remaining candidate with the least cumulative joint motion.
    }
    \vspace{-1em}
    \label{fig:ik}
\end{wrapfigure}

The reach stage can be instantiated from different high-level inputs. We first consider the default object-centric setting, in which the inputs are an object mesh $\mathcal{M}$, a desired object trajectory $\tau_o$, and a task description. The planner must determine where to contact the object and select an initial hand configuration that allows a feasible approach and downstream arm motion. Existing geometry-based grasp synthesis methods~\cite{chen2025bodex,wei2024dro} can generate locally stable candidates but do not consider task affordances or the full object trajectory. A candidate may therefore be functionally unsuitable, or it may be initially reachable but become infeasible later because of joint limits or collisions (Figure~\ref{fig:ik}). We address these issues through VLM-based affordance reasoning and future-aware IK selection to select a task-relevant initial configuration for which the induced arm motion remains feasible along $\tau_o$.

\noindent\textbf{VLM-based affordance reasoning.} 
Geometric stability alone does not determine whether a grasp is suitable for the task. For example, grasping a drill by its chuck may be stable but does not support functional tool use. We therefore use a VLM to identify a task-relevant contact region. We render the object from four viewpoints and prompt a VLM to identify the task-relevant grasping part and predict keypoints spanning its extent. The selected 2D keypoints are back-projected onto the mesh and aggregated across views by voting, yielding a task-relevant contact region $\mathcal{R} \subset \partial\mathcal{M}$.

\noindent\textbf{Region-conditioned candidate generation.} 
We use BODex's lightweight initialization stage~\cite{chen2025bodex} to sample candidate hand poses and joint configurations from object geometry within $\mathcal{R}$, yielding $\mathcal{G} = \{g_i\}_{i=1}^{N}$ ($N=10$).
Each $g_i$ is an approximate initialization rather than a final grasp. The learned contact-acquisition primitive then establishes contact more robustly, as shown in our experiments.

\noindent \textbf{Trajectory-aware selection.} For each $g_i$, we first plan a collision-free approach trajectory $\xi_{\mathrm{app}}$ from the robot's current configuration to the candidate initialization. We then roll out the nominal hand motion induced by keeping the hand-object transform fixed while the object follows $\tau_o$. We then solve inverse kinematics (IK) along this motion and discard candidates for which the approach is infeasible, the IK solutions violate joint limits, or the solver fails to converge. Among the remaining candidates, we select $g^\star$ with the least cumulative joint motion:
  $
  g^\star =
  \arg\min_{g_i\in\mathcal{G}_{\mathrm{valid}}}
  \sum_{t=1}^{T}
  \left\|q^{\mathrm{arm}}_{i,t} - q^{\mathrm{arm}}_{i,t-1}\right\|_2,
  $
  where $\mathcal{G}_{\mathrm{valid}}$ is the set of remaining candidates and $q^{\mathrm{arm}}_{i,t}$ is the arm IK solution for candidate $g_i$ at timestep $t$.
Implementation details are provided in the Appendix~\ref{sec:vlm-select}.

\subsection{Reach Specifications from Human Guidance and External Models}
\label{sec:inference}

\begin{figure}[t]
    \centering
    \includegraphics[width=\textwidth]{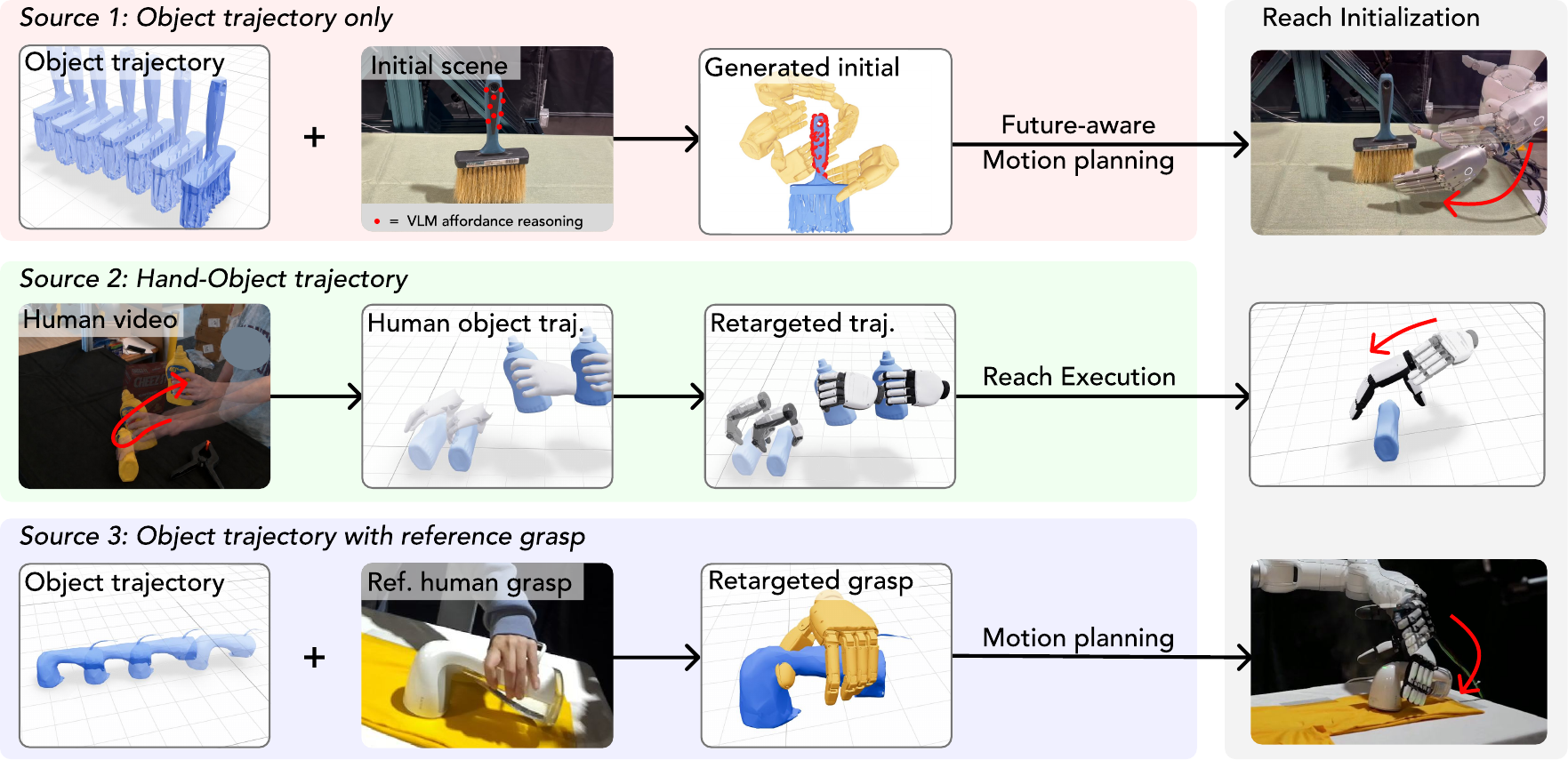}
    \caption{\small\textbf{Alternative sources of reach specifications.} Given a desired object trajectory, \ours can construct a reach specification through future-aware planning or use additional guidance from reference grasps or human approach trajectories. The shared contact-acquisition primitive refines the resulting approximate initializations into stable contact.}
    \vspace{-1em}
    \label{fig:input_sources}
\end{figure}

Section~\ref{sec:planning} constructs the reach specification $\rho$ from object-side information. Alternatively, human demonstrations or external models can supply contact regions, initial hand configurations, or approach trajectories directly, replacing the corresponding planning steps (Figure~\ref{fig:input_sources}). The resulting reach specification is executed by the same contact-acquisition primitive and trajectory tracker.

\paragraphc{Contact regions or initial configurations.}
External models or human demonstrations can provide a task-relevant contact region or an initial hand configuration. Such a contact region replaces VLM-based affordance reasoning, while an initial configuration also bypasses candidate generation. The remaining planning steps construct a feasible approach and check downstream arm feasibility to form $\rho$.

\paragraphc{Retargeted human approaches.}
Human-object datasets such as DexYCB~\cite{chao2021dexycb} and OakInk~\cite{yang2022oakink} provide synchronized hand and object motions. We retarget the human approach segment to the robot hand to obtain $\xi_{\mathrm{app}}$ and its terminal configuration $g_{\mathrm{init}}$, which together define $\rho$. This supplied approach replaces approach generation. Its endpoint need not be an exact final grasp: the contact-acquisition primitive establishes stable contact, after which the trajectory tracker follows $\tau_o$.

Richer guidance replaces more planning steps, but may not account for robot-specific reachability or collisions. We therefore retain future-aware planning as the default for arm-hand execution. Qualitative examples of these alternative compositions are available on the project website.

\subsection{Execution from High-Level Specifications}
\label{sec:execution}

Regardless of the high-level source, the execution backend receives a desired object trajectory $\tau_o$ and a reach specification $\rho$. During real-world deployment, we use SAM2~\cite{ravi2024sam2} to segment the target object and FoundationPose~\cite{wen2024foundationpose} to estimate and track its pose. \ours executes the task in three stages: \emph{reach, acquire, and move}.

\paragraphc{Reach.}
The robot follows the approach trajectory specified by $\rho$. The system switches to the contact-acquisition primitive once the minimum distance between the hand links and the object surface falls below $\epsilon_{\mathrm{approach}}$.

\paragraphc{Contact acquisition.}
At this transition, the contact-acquisition primitive takes control of the wrist and fingers. It locally adjusts the hand to compensate for object pose errors, geometry mismatch, and contact uncertainty. During this phase, we run the proprioception-based contact predictor described in Section~\ref{sec:rl}. Once the predictor detects stable contact, the system proceeds to the move stage.

\paragraphc{Pose-conditioned end-effector tracking.}
After stable contact is detected, the robot performs a short lift before tracking the desired object trajectory. We approximate the hand-object transform as fixed during tracking. At each timestep, we compute an end-effector target from the desired object pose and the estimated hand-object transform, correct the target using the observed object pose error, and send it to the low-level position controller.

\section{Experiments}

We evaluate the contact-acquisition primitive's robustness, cross-embodiment transfer, and contribution to end-to-end task success. We further examine its reuse across workspaces and with different high-level sources, and evaluate key design choices through ablations.

\begin{wrapfigure}{r}{0.6\textwidth}
    \centering
    \vspace{-0.5em}
    \includegraphics[width=\linewidth]{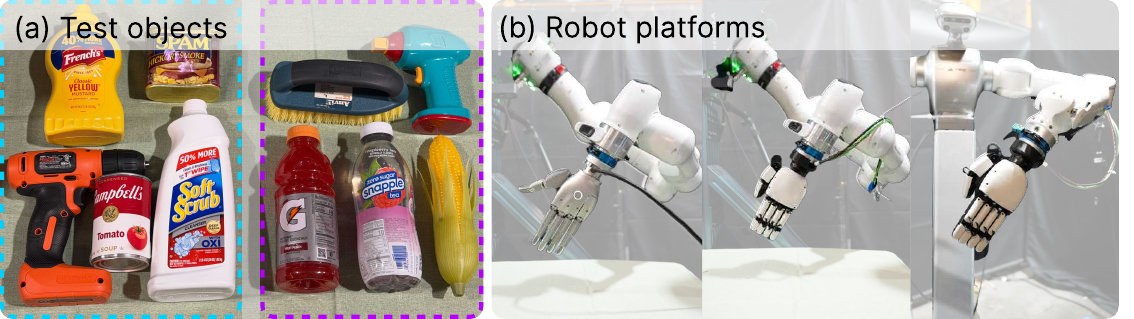}
    \caption{\small \textbf{Real-world evaluation setup.} (a) in-distribution (blue) and out-of-distribution (purple) test objects; (b) three robot embodiments used for evaluation.}
    \vspace{-1.5em}
    \label{fig:real_setup}
\end{wrapfigure}

\paragraphc{Hardware setup.}
We evaluate \ours on multiple arm-hand platforms, including Franka Research 3 (7-DoF) and Tianji (7-DoF) arms paired with XHand (12-DoF) and Sharpa Wave (22-DoF) hands, as shown in Figure~\ref{fig:real_setup}.

\paragraphc{Object assets and perception stack.}
We obtain object meshes using AR Code\footnote{\vspace{-2mm}https://ar-code.com/} on an iPhone and use them for planning and pose tracking. During execution, SAM2~\cite{ravi2024sam2} segments the target object in the initial frame, and FoundationPose~\cite{wen2024foundationpose} estimates and tracks its 6D pose using the mesh and segmentation mask. FoundationPose runs at 45\,Hz, while the policy runs at 10\,Hz and uses the latest pose estimate at each control step.

\begin{table}[tb]
\centering
\caption{\small\textbf{Simulation lifting success on three left hands (\%).} BODex and DRO are executed as open-loop grasp baselines. CrossDex and \ours are evaluated as specialist policies trained separately for each hand and unified policies trained jointly across all three hands. Learned-policy results report mean $\pm$ standard deviation across three seeds. Best results per column are bolded.}
\setlength{\tabcolsep}{6pt}
\renewcommand{\arraystretch}{1.1}
\resizebox{0.78\linewidth}{!}{
\begin{tabular}{llcccc}
\toprule
Method & Setting & XHand & Sharpa & Allegro & Avg. \\
\midrule
BODex~\cite{chen2025bodex} & - & 45.4 & 27.2 & 43.6 & 38.7 \\
DRO~\cite{wei2024dro} & - & 31.3 & 47.0 & 43.2 & 40.5 \\
\midrule
\multirow{2}{*}{CrossDex~\cite{yuan2025cross}}
& Specialist
& 30.4 $\pm$ 10.9
& 74.9 $\pm$ 1.5
& 34.5 $\pm$ 3.3
& 46.6 $\pm$ 4.1 \\
& Unified
& 44.7 $\pm$ 2.9
& 52.8 $\pm$ 3.7
& 57.6 $\pm$ 1.0
& 51.7 $\pm$ 1.5 \\
\midrule

\multirow{2}{*}{\textbf{Ours}}
& Specialist
& 64.3 $\pm$ 0.7
& \textbf{81.4 $\pm$ 1.2}
& 80.4 $\pm$ 1.8
& 75.4 $\pm$ 1.0 \\
& Unified
& \textbf{74.1 $\pm$ 1.2}
& 75.0 $\pm$ 1.1
& \textbf{87.1 $\pm$ 4.4}
& \textbf{78.7 $\pm$ 1.1} \\
\bottomrule
\end{tabular}
}
\vspace{-1em}
\label{tab:main}
\end{table}

\paragraphc{Simulation setup.}
We train a unified policy for XHand, Sharpa, and Allegro in IsaacLab~\cite{mittal2025isaac} using PPO~\cite{schulman2017proximal}. We train for 2,000 epochs using 4,096 parallel environments. The training set includes 16 YCB objects~\cite{calli2017yale}, with 4,000 initial configurations generated using BODex for each object–resting-pose pair.

\subsection{Is It a Reliable General-Purpose Contact Primitive?}

We isolate the low-level contact-acquisition primitive by evaluating object lifting in simulation and the real world, without full-arm planning.

\subsubsection{Simulation grasping evaluation}

We use a floating-hand tabletop setup and evaluate on DexGraspNet objects~\cite{wang2022dexgraspnet}, following the preprocessing protocol of BODex~\cite{chen2025bodex}. After filtering out flat objects and rescaling the remaining objects to four sizes, the evaluation set contains 9,588 assets and 54,149 object-scale-pose instances. We run simulations in IsaacLab with the friction coefficient set to 1.5 and the object mass set to 100\,g. All physics parameters are fixed across evaluations.

\paragraphc{Baselines.}
We compare against two representative grasp synthesis methods. BODex~\cite{chen2025bodex} optimizes grasp poses for each robot-object pair using bilevel optimization. DRO~\cite{wei2024dro} is a cross-embodiment grasp prediction network, for which we use the authors' released checkpoints. Since both methods output grasp targets rather than closed-loop acquisition policies, we execute their outputs open-loop using their original execution controllers.

\paragraphc{Evaluation protocol.}
For each object-scale-pose instance, we generate 20 initial poses using BODex's first stage (Section~\ref{sec:planning}) and share them across methods. BODex refines them via bilevel optimization, DRO uses them to specify initial approach directions, and our policy starts directly from them for closed-loop contact acquisition. In the same tabletop setup, BODex and DRO execute open-loop from predicted pre-grasps, effectively assuming successful motion planning to these poses. Our method additionally handles the approach from the raw initial poses.

\paragraphc{Metrics.} We report lifting success rates across hand embodiments, objects, and initial object poses. A trial is successful if the hand establishes contact, lifts the object by at least 5\,cm, and keeps it stable for the full evaluation horizon.

\paragraphc{Results.}
The results are shown in Table~\ref{tab:main}. \ours consistently outperforms BODex and DRO in lifting success across all three hands.
Notably, although the policy is trained on only 16 YCB objects, it generalizes to the full evaluation set of 9,588 DexGraspNet assets. 

The success rates for BODex and DRO are lower than those reported in their original papers due to our stricter tabletop protocol. Prior work initializes objects without gravity and applies gravity only after contact is established, whereas we place objects on a table with gravity enabled from the start and execute directly. Appendix \ref{sec:appendix-baseline-protocol} details the baseline implementations and evaluation protocols and presents controlled comparisons.

\subsubsection{Real-world grasping experiments}
We evaluate the contact-acquisition primitive using the same checkpoint on three arm-hand platforms: Franka-Sharpa, Franka-XHand, and Tianji-XHand. This tests both cross-hand generalization and transfer across arm platforms through the end-effector control interface. The evaluation setup is shown in Figure~\ref{fig:real_setup}. We run 5 trials on each of 10 objects, for 50 rollouts per platform. A trial is successful if the hand establishes contact, lifts the object by at least 5\,cm, and keeps it stable for at least 3 seconds. \ours achieves success rates of 90\%, 90\%, and 96\% on Franka-Sharpa, Franka-XHand, and Tianji-XHand, respectively. The slightly lower success rates on Franka are mainly due to the more constrained reachable workspace in our Franka setup.

\subsection{Does It Generalize across and beyond Training Hands?}

\subsubsection{Cross-hand evaluation and representation comparison}

The lower portion of Table~\ref{tab:main} compares \ours with CrossDex~\cite{yuan2025cross}, an RL baseline that uses a shared action space through human-hand retargeting. To isolate the effect of the observation–action representation, we reimplement CrossDex within our framework. We adopt the observation and action design from its released code while keeping the reward, PPO hyperparameters, and training and evaluation protocol the same as those used for \ours. Specialist policies are trained separately for each hand, while unified policies are trained jointly on XHand, Sharpa, and Allegro. Our unified policy with the per-link canonical representation achieves 78.7\% average success versus 51.7\% for CrossDex. Joint training also improves average success over our specialist policies, supporting positive transfer across morphologies.

\subsubsection{Zero-shot transfer to unseen Allegro variants}
\begin{table}[t]
  \centering
  \caption{\small \textbf{Zero-shot generalization to unseen Allegro variants.}
  Lifting success rate (\%), reported as mean $\pm$ standard deviation over three seeds.
  \texttt{xyzw} denotes per-finger link counts
  (\texttt{4444}\,=\,original Allegro).
  Oracle trains one policy per variant; Allegro-only trains on the original Allegro;
  Three-hand is our unified policy over XHand+Sharpa+original Allegro, evaluated zero-shot.}
  \label{tab:allegro_zeroshot}
  \setlength{\tabcolsep}{4pt}
\renewcommand{\arraystretch}{1.1}
\footnotesize
\resizebox{0.78\linewidth}{!}{
  \begin{tabular}{lcccc}
  \toprule
  Variant
  & \texttt{4433}
  & \texttt{4333}
  & \texttt{4324}
  & Avg. \\
  \midrule
  Oracle (per-variant)
  & 69.0 $\pm$ 6.8
  & 71.6 $\pm$ 1.1
  & \textbf{76.0 $\pm$ 4.0}
  & \textbf{72.2 $\pm$ 1.2} \\

  Allegro-only (zero-shot)
  & 29.4 $\pm$ 19.7
  & 12.9 $\pm$ 8.6
  & 9.4 $\pm$ 8.1
  & 17.2 $\pm$ 12.1 \\

  \textbf{Three-hand (zero-shot)}
  & \textbf{74.4 $\pm$ 6.5}
  & \textbf{72.2 $\pm$ 4.8}
  & 69.6 $\pm$ 9.6
  & 72.1 $\pm$ 6.9 \\
  \bottomrule
  \end{tabular}
  }
  \end{table}

\begin{figure}
\centering
\includegraphics[width=\linewidth]{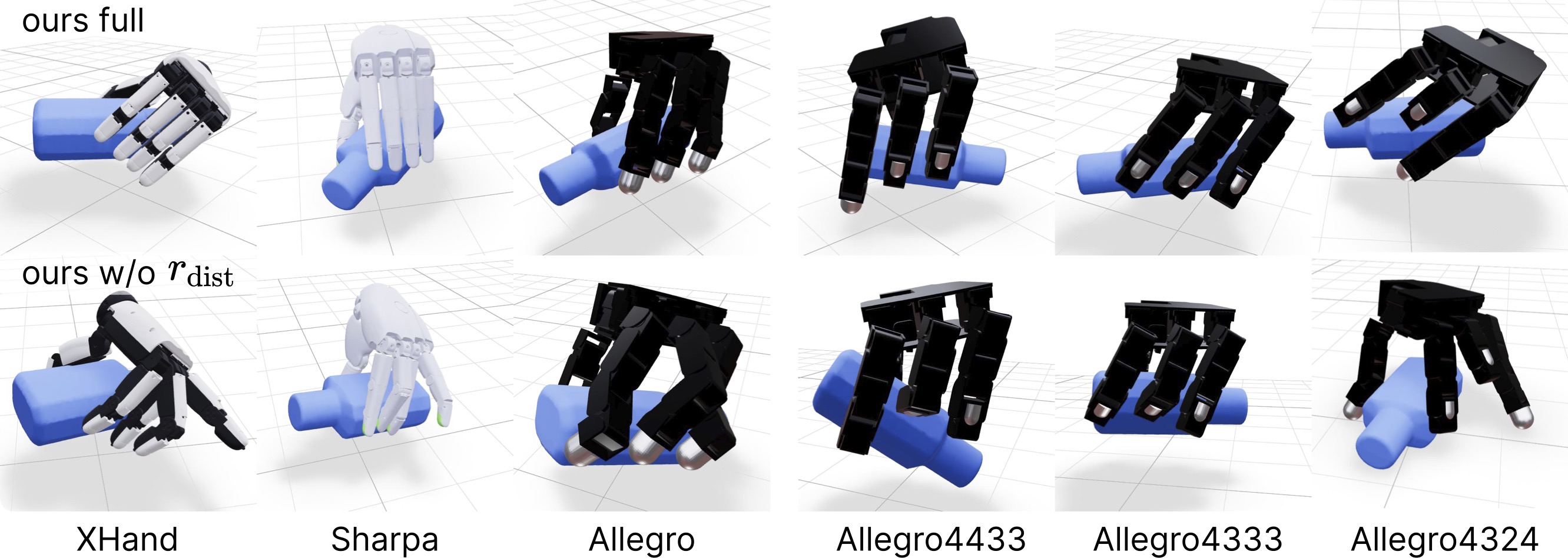}
\caption{\small\textbf{Qualitative grasps on training and unseen
embodiments.} We compare the full \ours (top) with the variant
without $r_{\mathrm{dist}}$ (bottom) on three training hands and three
unseen Allegro variants.}
\label{fig:grasp-vis}
\vspace{-1em}
\end{figure}

To further probe cross-embodiment generalization, we test unseen Allegro variants generated by removing finger links. We evaluate our three-hand unified policy, trained on XHand, Sharpa, and the original Allegro hand, \emph{zero-shot} on these variants. We compare it with (i) per-variant oracle policies trained directly on each variant and (ii) a policy trained only on the original Allegro hand. As shown in Table~\ref{tab:allegro_zeroshot}, our three-hand policy averages 72.1\% success across the three variants, matching the average of the per-variant oracle policies (72.2\%), whereas the Allegro-only policy achieves 17.2\%. Transfer therefore comes from multi-morphology training, not the canonical representation alone. Figure~\ref{fig:grasp-vis} further shows that the policy produces coordinated grasp configurations on these unseen variants.

\subsection{Does the Primitive Improve End-to-End Task Success?}

\begin{figure}[t]
    \centering
    \includegraphics[width=\columnwidth]{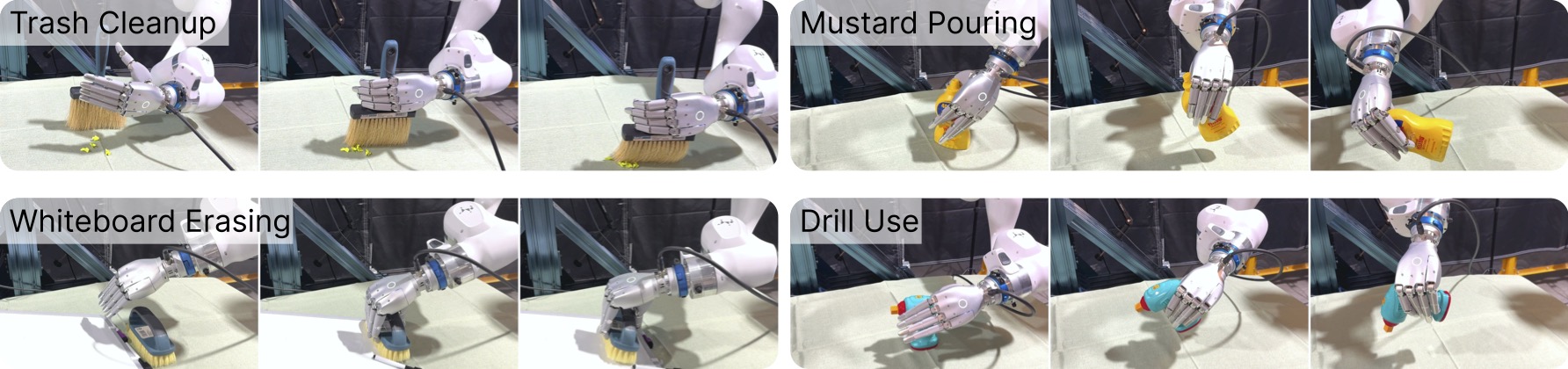}
    \caption{\small Qualitative illustration of four real-world tasks.}
    \label{fig:task}
\end{figure}

We evaluate whether the contact-acquisition primitive improves end-to-end execution on four real-world manipulation tasks using the Franka-Sharpa platform. For each task, the robot must approach the object, acquire stable contact, and move the object along the specified trajectory.
We compare \ours with two matched open-loop variants. Each replaces our region-conditioned candidate generation with the full grasp generation procedure of BODex~\cite{chen2025bodex} or DRO~\cite{wei2024dro} and executes the predicted grasp open-loop. All other components, including perception, VLM-based affordance reasoning, future-aware IK selection, and trajectory tracking, are kept fixed, isolating the contribution of closed-loop contact acquisition.

\paragraphc{Real-world tasks.}
We evaluate \textit{Trash Cleanup} (grasp a brush and sweep trash), \textit{Mustard Pouring} (grasp a mustard bottle and rotate it to a pouring orientation), \textit{Whiteboard Erasing} (grasp an eraser and sweep across the board), and \textit{Drill Use} (grasp a power drill and rotate it to a working orientation), as illustrated in Figure~\ref{fig:task}.
These tasks require both translation and rotation, and some require functional contacts, such as grasping the drill near its handle. Each task is repeated 10 times across varied initial poses.

\begin{table}[t]
\centering
\caption{\small \textbf{Real-world task results.} Per-task grasp success rate
(gSR) and task success rate (tSR) across four manipulation
tasks on Franka-Sharpa platform, 10 trials each. Best per column in bold.}
\label{tab:real_tasks}
\setlength{\tabcolsep}{6pt}
\renewcommand{\arraystretch}{1.15}
\resizebox{\columnwidth}{!}{%
\begin{tabular}{lcccccccccc}
\toprule
& \multicolumn{2}{c}{\texttt{Trash Cleanup}}
& \multicolumn{2}{c}{\texttt{Mustard Pouring}}
& \multicolumn{2}{c}{\texttt{Whiteboard Erasing}}
& \multicolumn{2}{c}{\texttt{Drill Use}}
& \multicolumn{2}{c}{Avg. (\%)} \\
\cmidrule(lr){2-3} \cmidrule(lr){4-5} \cmidrule(lr){6-7} \cmidrule(lr){8-9} \cmidrule(lr){10-11}
Method & gSR & tSR & gSR & tSR & gSR & tSR & gSR & tSR & gSR & tSR \\
\midrule
DRO   & 6/10 & 0/10  & 5/10 & 5/10 & 5/10 & 3/10  & 7/10 & 6/10 & 57.5 & 35.0 \\
BODex & 5/10 & 2/10  & 4/10 & 2/10 & 7/10 & 3/10  & 7/10 & 7/10 & 57.5 & 35.0 \\
Ours  & \textbf{8/10} & \textbf{7/10} & \textbf{8/10} & \textbf{7/10} & \textbf{10/10} & \textbf{9/10} & \textbf{9/10} & \textbf{9/10} & \textbf{87.5} & \textbf{80.0} \\
\bottomrule
\end{tabular}%
}
\vspace{-1.em}
\end{table}

\paragraphc{Metrics.}
We report grasp success rate and task success rate. Grasp success measures whether the robot acquires stable contact, while task success measures whether it completes the downstream trajectory-tracking task.

\paragraphc{Results.}
Table~\ref{tab:real_tasks} shows that \ours achieves the highest grasp and task success rates on every task, with an average task success rate of 80.0\% versus 35.0\% for both baselines. Baseline failures include unsuitable contact regions, object displacement during approach, and cage-like grasps lacking firm opposing contact between the fingers and thumb. The last failure mode explains DRO's 0/10 task success on \textit{Trash Cleanup}: loose contact around the brush handle causes the brush to slip during sweeping. While the open-loop baselines are sensitive to planning errors, our closed-loop controller adjusts the wrist and fingers to recover from imperfect initializations.

\subsection{Does Decomposition Improve Robustness to Workspace Variations?}

SimToolReal~\cite{kedia2026simtoolreal} and Dex4D~\cite{kuang2026dex4d} jointly learn contact acquisition and object tracking with embodiment-specific policies. In contrast, \ours uses a reusable contact primitive, arm-specific IK, and object-pose-conditioned tracking to separate contact acquisition from robot-specific reachability.

\begin{wrapfigure}{r}{0.45\textwidth}
\centering
\includegraphics[width=\linewidth]{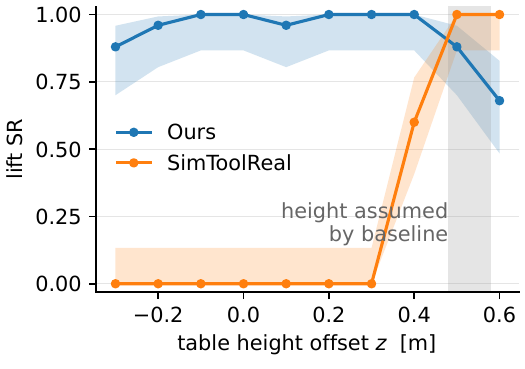}
\caption{ Success rate versus table height offset $z$ on the SimToolReal eraser object.}
\label{fig:height_profile}
\vspace{-1.em}
\end{wrapfigure}

We evaluate \ours on the eraser object from SimToolReal's DexToolBench, using a $5\,$cm lifting trajectory, over 600 simulated episodes. We vary the object's planar position and the table's height offset $z$ relative to the robot base (Figure~\ref{fig:height_profile}).

Near SimToolReal's training configuration, its success rate is 0.983, compared with 0.792 for \ours. Our failures are mainly due to IK singularities under a constrained end-effector orientation. Below $z=0.4\,$m, SimToolReal's success rate drops to 0.002, while \ours achieves 0.914.
In this comparison, we reuse our contact primitive on SimToolReal's iiwa-Sharpa platform and evaluate SimToolReal using its released checkpoint. Adapting SimToolReal to a different arm embodiment would instead require roughly seven days of training, whereas our primitive is arm-agnostic by design. These results support separating workspace-specific reachability from reusable contact control.

\subsection{Does It Support High-Level Specifications from Different Sources?}

We test whether \ours accepts high-level specifications from diverse sources. These include object trajectories obtained from different sources and unseen human demonstrations that also specify the hand's approach. Across all sources, the contact-acquisition primitive remains fixed, without task- or trajectory-specific fine-tuning.

\subsubsection{Object trajectories from diverse sources} 

\begin{wrapfigure}{r}{0.54\textwidth}
    \centering
    \includegraphics[width=\linewidth]{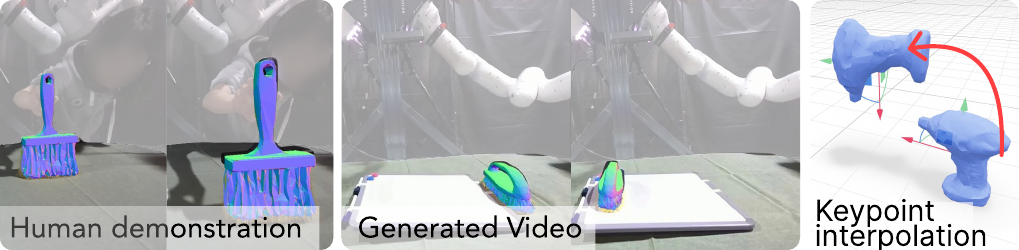}
    \caption{\small Visualization of three sources of object trajectories.}
    \vspace{-1.5em}
    \label{fig:trajectory}
\end{wrapfigure}

Figure~\ref{fig:trajectory} shows that object trajectories can be obtained from human demonstrations, generated videos, and manually specified object-pose keypoints. For human demonstrations, we track object SE(3) trajectories using FoundationPose~\cite{wen2024foundationpose}. For generated videos, we follow RigVid~\cite{patel2025robotic}, estimating per-frame depth with RollingDepth~\cite{ke2025video} and recovering object trajectories with FoundationPose~\cite{wen2024foundationpose}. For keypoints, we interpolate between user-specified object poses. The resulting trajectories from all three sources can be used directly by the \ours pipeline.

\subsubsection{Zero-shot execution of unseen human demonstrations}

We evaluate all 499 left-hand DexYCB sequences~\cite{chao2021dexycb}, each containing human-hand motion and an object trajectory. We kinematically retarget the hand motion to XHand for the approach, acquire stable contact with our policy, and then track the object trajectory. We use the same policy for all sequences without fine-tuning.
We compare against kinematic replay, which directly executes the retargeted trajectories. Neither kinematic replay nor \ours performs per-demonstration test-time optimization. We additionally report SPIDER~\cite{pan2025spider}, which generates robot controls via per-demonstration physics-informed optimization, as a reference with an additional test-time optimization stage.
We use SPIDER's task-success criterion and released MuJoCo implementation. Our reimplementations of SPIDER in IsaacLab and ManiSkill failed to achieve stable execution, so we omit their results. Our policy is trained in IsaacLab and evaluated zero-shot in ManiSkill and MuJoCo to assess sim-to-sim transfer.

\begin{figure}
    \centering
    \includegraphics[width=\linewidth]{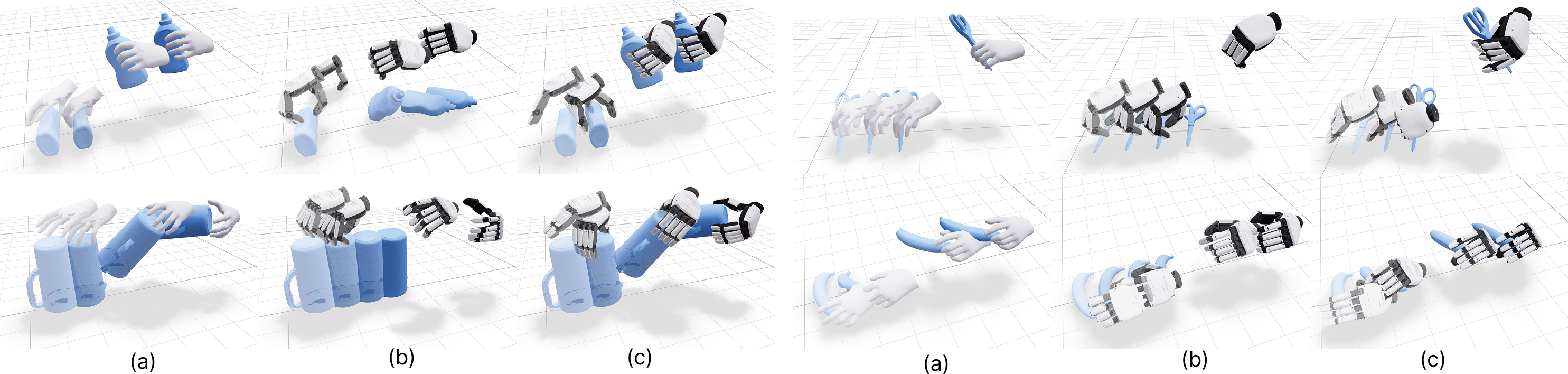}
    \caption{\small \textbf{Execution of unseen DexYCB demonstrations.} Transparent hands indicate motion over time. (a) Annotated human hand-object motion. (b) Open-loop kinematic replay fails to maintain contact in physics simulation. (c) Our reactive primitive acquires stable contact and executes the demonstrated object trajectory.}
    \label{fig:human-to-robot}
\end{figure}

On these 499 demonstrations, kinematic replay achieves a success rate of 12.6\% in MuJoCo, whereas \ours achieves 71.5\% in IsaacLab, 72.9\% in ManiSkill, and 69.7\% in MuJoCo, demonstrating sim-to-sim transfer. SPIDER achieves 76.0\% in MuJoCo but requires per-demonstration optimization and reoptimization whenever nominal object states or trajectories change. Figure~\ref{fig:human-to-robot} compares human demonstrations, kinematic replay, and executions by \ours.
  
\begin{table}[t]
\centering
\caption{\small \textbf{Ablation results.} Success rate (\%)
on three left hands.}
\label{tab:ablation}
\begin{tabular}{lcccc}
  \toprule
  Variant & XHand & Sharpa & Allegro & Avg. \\
  \midrule
  \textbf{Full \ours{}}
   & \textbf{74.1 $\pm$ 1.2}
   & 75.0 $\pm$ 1.1
   & \textbf{87.1 $\pm$ 4.4}
   & \textbf{78.7 $\pm$ 1.1} \\

  \midrule

  \quad w/o link position obs.
   & 49.4 $\pm$ 6.4
   & 57.9 $\pm$ 10.3
   & 75.7 $\pm$ 10.6
   & 61.0 $\pm$ 4.4 \\

  \quad proprio history length $=1$
   & 56.5 $\pm$ 5.1
   & 65.0 $\pm$ 2.1
   & 67.5 $\pm$ 3.0
   & 63.0 $\pm$ 3.3 \\

   \quad OHRA-style~\cite{wei2026one} obs. & 24.9 $\pm$ 22.8 & 31.2 $\pm$ 25.2 & 20.9 $\pm$ 15.6 & 25.7 $\pm$ 18.3 \\
   
   \midrule

  \quad w/o $r_{\mathrm{contact}}$
   & 0.0 $\pm$ 0.0
   & 0.0 $\pm$ 0.0
   & 0.0 $\pm$ 0.0
   & 0.0 $\pm$ 0.0 \\

  \quad w/o $r_{\mathrm{lift}}$
   & 0.8 $\pm$ 0.1
   & 2.7 $\pm$ 1.3
   & 1.0 $\pm$ 1.2
   & 1.5 $\pm$ 0.8 \\

  \quad w/o $r_{\mathrm{dist}}$
   & 70.6 $\pm$ 3.0
   & \textbf{76.9 $\pm$ 4.2}
   & 80.6 $\pm$ 7.0
   & 76.0 $\pm$ 1.6 \\
   
  \bottomrule
  \end{tabular}

\vspace{-1em}
\end{table}

\subsection{Ablations and Analysis}

We analyze two aspects of \ours: the primitive's observation and reward design (Section~\ref{sec:rl})
and the role of affordance reasoning and trajectory-aware selection in generating functional, feasible high-level specifications (Section~\ref{sec:planning}).
  
\subsubsection{Contact primitive design} 
\label{sec:ablation}
We vary one observation or reward setting at a time, holding all others fixed (Table~\ref{tab:ablation}). Removing link-position observations degrades success across hands. Reducing the proprioceptive history to one timestep removes temporal contact cues, lowering average success by 15.7 percentage points. We also compare with an OHRA-style~\cite{wei2026one} observation representation. In its original form, OHRA encodes only the object point cloud with a heavyweight point-cloud encoder, concatenates the resulting feature with a morphology VAE embedding, and uses the combined representation to predict grasps offline. This encoder is too large to train jointly with PPO at our training scale, so we adapt the representation to RL by (i) replacing the encoder with a lighter PointNet and (ii) adding the hand point cloud as input, while preserving OHRA's morphology VAE conditioning. Even after this adaptation, average success drops to 25.7\%, suggesting that per-link contact geometry provides a better representation for a closed-loop primitive.

Removing either $r_{\mathrm{contact}}$ or $r_{\mathrm{lift}}$ reduces average success to below 2\%, confirming their necessity. While $r_{\mathrm{dist}}$ modestly improves lifting on training hands on average, it suppresses finger inversion and distorted joint bending (Figure~\ref{fig:grasp-vis}). Under morphology shift, removing this regularization lowers average success on three unseen Allegro variants from 72.1\% to 55.1\%.

\subsubsection{Affordance- and Future-Aware Grasp Selection}
\label{sec:vlm_eval}

\begin{table}[t]
\centering
\caption{\small \textbf{Affordance- and Future-aware selection on \texttt{Drill Use} (20 scenes).}  Random and IK-only use affordance-agnostic candidates and select by random sampling and trajectory feasibility, respectively. VLM variants first generate candidates from a task-relevant contact region and then apply future-aware IK selection.}
\label{tab:vlm_selection}
\renewcommand{\arraystretch}{1.1}
\footnotesize
\resizebox{\linewidth}{!}{
\begin{tabular}{lcccccc}
\toprule
Selection  & Affordance & Future-aware                   & gSR (\%) & tjSR (\%) & func. (\%) & tSR (\%) \\
\midrule
Random    & \xmark & \xmark             & \textbf{100}           & 80            & 35            & 20 \\
IK-only   & \xmark & \cmark                   & 95           & \textbf{90}            & 30            & 30 \\
VLM (gpt-5.6-sol) & \cmark & \cmark         & 90           & 70            & 80            & 55 \\
VLM (gemini-flash-latest) & \cmark & \cmark    & 70 & 50   & 65            & 50 \\
\textbf{VLM (claude-opus-4-7)} & \cmark & \cmark & 80 & 70            & \textbf{85}            & \textbf{65} \\
\textbf{VLM (claude-opus-5)} & \cmark & \cmark & 95 & 75           & \textbf{85}            & 60 \\
\bottomrule
\end{tabular}
}
\vspace{-2em}
\end{table}
We evaluate affordance-conditioned candidate generation and future-aware selection on 20 matched \textit{Drill Use} scenes, which require grasping the handle instead of the chuck. All methods use the same scene initializations, with position randomized by $\pm0.10\,$m along $x$ and $y$ and orientation by $\pm30^\circ$. 
Table~\ref{tab:vlm_selection} reports grasp success rate (gSR), task trajectory success rate (tjSR), functional correctness (func.), and overall task success rate (tSR). Trajectory success requires the drill to rotate by $45^\circ$ about the specified axis while remaining grasped by the robot. Overall task success requires grasp success, trajectory success, and an affordance-correct grasp to all hold in the same trial. All rates are computed over the same 20 scenes. Scenes with no valid candidate terminate without execution and count as failures for all metrics.

Random selection achieves 100\% gSR but only 20\% tSR: an initially valid grasp may not support the task. Future-aware IK selection raises tjSR from 80\% to 90\%, 
but functional correctness and overall task success remain at 30\%: successful trajectory execution alone does not ensure a task-appropriate grasp.
Across VLM backbones, affordance conditioning improves func. to 65--85\% and tSR to 50--65\%. Restricting candidates to task-relevant regions lowers tjSR to 50--75\% but improves overall task success.

\section{Conclusion}
\label{sec:conclusion}

We presented \ours, a \textit{reach-acquire-move} framework for dexterous manipulation from object trajectories and flexible reach specifications. At its core, a cross-embodiment contact-acquisition primitive establishes stable contact from approximate reach specifications, decoupling local contact control from arm-specific reaching and downstream tracking. Simulation and real-world experiments demonstrate the primitive's reuse across objects, embodiments, tasks, workspaces, and high-level input sources.

\paragraphc{Limitations and future work.}
First, high-level planning and contact acquisition operate sequentially. Tighter feedback between them could enable online recovery from poor reach specifications.
Second, trajectory tracking relies on visual object pose estimation, which can fail under severe hand-object occlusion. Incorporating tactile sensing or visuotactile state estimation is a natural next step.
Finally, we treat the hand-object transform as approximately fixed after grasping. Extending \ours to handle changes in this transform would enable more precise in-hand manipulation tasks, such as unscrewing.

\acknowledgments{We thank Tingfan Wu for his help with hardware support. We are grateful to Shuo Cheng for proposing to use a vision-language model to query task-relevant keypoints, as well as for insightful discussions and suggestions on VLM-based affordance reasoning. We thank Qiyang Yan for substantially accelerating the FoundationPose tracking speed, and Jiayuan Mao and Ruoshi Liu for helpful discussions and suggestions.}

\bibliographystyle{plainnat}
\bibliography{references}

@inproceedings{feng2026regrind,
  title={A Minimalist Retargeting-Guided Reinforcement Learning Recipe for Dexterous Manipulation},
  author={Feng, Yunhai and Leung, Natalie and Wang, Jiaxuan and Yang, Lujie and Qi, Haozhi and Culbertson, Preston},
  booktitle={CoRL},
  year={2026}
}

@inproceedings{qi2023rotateit,
  title={General In-Hand Object Rotation with Vision and Touch},
  author={Qi, Haozhi and Yi, Brent and Suresh, Sudharshan and Lambeta, Mike and Ma, Yi and Calandra, Roberto and Malik, Jitendra},
  booktitle={CoRL},
  year={2023}
}

@inproceedings{xu2023unidexgrasp,
  title={{UniDexGrasp}: Universal Robotic Dexterous Grasping via Learning Diverse Proposal Generation and Goal-Conditioned Policy},
  author={Xu, Yinzhen and Wan, Weikang and Zhang, Jialiang and Liu, Haoran and Shan, Zikang and Shen, Hao and Wang, Ruicheng and Geng, Haoran and Weng, Yijia and Chen, Jiayi and Liu, Tengyu and Yi, Li and Wang, He},
  booktitle={CVPR},
  year={2023}
}

@inproceedings{wan2023unidexgrasppp,
  title={{UniDexGrasp++}: Improving Dexterous Grasping Policy Learning via Geometry-aware Curriculum and Iterative Generalist-Specialist Learning},
  author={Wan, Weikang and Geng, Haoran and Liu, Yun and Shan, Zikang and Yang, Yaodong and Yi, Li and Wang, He},
  booktitle={ICCV},
  year={2023}
}

@inproceedings{zhang2024graspxl,
  title={{GraspXL}: Generating Grasping Motions for Diverse Objects at Scale},
  author={Zhang, Hui and Christen, Sammy and Fan, Zicong and Hilliges, Otmar and Song, Jie},
  booktitle={ECCV},
  year={2024}
}

@article{zhang2026congraspxl,
  title={{ConGraspXL}: Controllable Constraint-Conditioned Dexterous Grasping Motion Synthesis},
  author={Zhang, Hui and Meboldt, Mirko and Song, Jie},
  journal={arXiv:2609.16319},
  year={2026}
}

@inproceedings{agarwal2023functional,
  title={Dexterous Functional Grasping},
  author={Agarwal, Ananye and Uppal, Shagun and Shaw, Kenneth and Pathak, Deepak},
  booktitle={CoRL},
  year={2023}
}

@article{huang2025fungrasp,
  title={{FunGrasp}: Functional Grasping for Diverse Dexterous Hands},
  author={Huang, Linyi and Zhang, Hui and Wu, Zijian and Christen, Sammy and Song, Jie},
  journal={RA-L},
  year={2025}
}

@inproceedings{wei2025afforddexgrasp,
  title={{AffordDexGrasp}: Open-Set Language-Guided Dexterous Grasp with Generalizable-Instructive Affordance},
  author={Wei, Yi-Lin and Lin, Mu and Lin, Yuhao and Jiang, Jian-Jian and Wu, Xiao-Ming and Zeng, Ling-An and Zheng, Wei-Shi},
  booktitle={ICCV},
  year={2025}
}

@inproceedings{patel2025getzero,
  title={{GET-Zero}: Graph Embodiment Transformer for Zero-shot Embodiment Generalization},
  author={Patel, Austin and Song, Shuran},
  booktitle={ICRA},
  year={2025}
}

@inproceedings{adalibieke2026adadextrack,
  title={{AdaDexTrack}: Dynamic Modulation for Adaptive and Generalizable Dexterous Manipulation Tracking},
  author={Adalibieke, Jianibieke and Han, Qianwei and Liu, Xueyi and Qin, Yuzhe and Yi, Li},
  booktitle={CVPR},
  year={2026}
}

@article{gupta2026grasp2act,
  title={Grasp to Act: Dexterous Grasping for Tool Use in Dynamic Settings},
  author={Gupta, Harsh and Mirzaee, Mohammad Amin and Yuan, Wenzhen},
  journal={RA-L},
  year={2026}
}

@inproceedings{christen2022dgrasp,
  title={{D-Grasp}: Physically Plausible Dynamic Grasp Synthesis for Hand-Object Interactions},
  author={Christen, Sammy and Kocabas, Muhammed and Aksan, Emre and Hwangbo, Jemin and Song, Jie and Hilliges, Otmar},
  booktitle={CVPR},
  year={2022}
}

@inproceedings{dalal2024local,
  title={Local Policies Enable Zero-shot Long-horizon Manipulation},
  author={Dalal, Murtaza and Liu, Min and Talbott, Walter and Chen, Chen and Pathak, Deepak and Zhang, Jian and Salakhutdinov, Ruslan},
  booktitle={ICRA},
  year={2025}
}

@inproceedings{hsu2024spot,
  title={{SPOT}: {SE(3)} Pose Trajectory Diffusion for Object-Centric Manipulation},
  author={Hsu, Cheng-Chun and Wen, Bowen and Xu, Jie and Narang, Yashraj and Wang, Xiaolong and Zhu, Yuke and Biswas, Joydeep and Birchfield, Stan},
  booktitle={ICRA},
  year={2025}
}

@inproceedings{gupta2026lucid,
  title={{LUCID}: Learning Embodiment-Agnostic Intent Models from Unstructured Human Videos for Scalable Dexterous Robot Skill Acquisition},
  author={Gupta, Harsh and Shi, Guanya and Yuan, Wenzhen},
  booktitle={CoRL},
  year={2026}
}

@inproceedings{patel2025robotic,
  title={Robotic Manipulation by Imitating Generated Videos Without Physical Demonstrations},
  author={Shivansh Patel and Shraddhaa Mohan and Hanlin Mai and Unnat Jain and Svetlana Lazebnik and Yunzhu Li},
  booktitle={ICLR},
  year={2026},
}

@inproceedings{chen2024object,
  title={Object-Centric Dexterous Manipulation from Human Motion Data},
  author={Yuanpei Chen and Chen Wang and Yaodong Yang and C. Karen Liu},
  booktitle={CoRL},
  year={2024},
}

@inproceedings{qi2022hand,
  author={Qi, Haozhi and Kumar, Ashish and Calandra, Roberto and Ma, Yi and Malik, Jitendra},
  title={In-Hand Object Rotation via Rapid Motor Adaptation},
  booktitle={CoRL},
  year={2022}
}

@article{chen2023visual,
  author={Tao Chen and Megha Tippur and Siyang Wu and Vikash Kumar and Edward Adelson and Pulkit Agrawal},
  title={Visual dexterity: In-hand reorientation of novel and complex object shapes},
  journal={Science Robotics},
  year = {2023}
}

@inproceedings{nvidia2022dextreme,
  author={Handa, Ankur and Allshire, Arthur and Makoviychuk, Viktor and Petrenko, Aleksei and Singh, Ritvik and Liu, Jingzhou and Makoviichuk, Denys and Van Wyk, Karl and Zhurkevich, Alexander and Sundaralingam, Balakumar and Narang, Yashraj and Lafleche, Jean-Francois and Fox, Dieter and State, Gavriel},
  title={DeXtreme: Transfer of Agile In-Hand Manipulation from Simulation to Reality},
  booktitle={ICRA},
  year={2023},
}

@inproceedings{pan2025spider,
  title={SPIDER: Scalable Physics-Informed Dexterous Retargeting},
  author={Chaoyi Pan and Changhao Wang and Haozhi Qi and Zixi Liu and Homanga Bharadhwaj and Akash Sharma and Tingfan Wu and Guanya Shi and Jitendra Malik and Francois Hogan},
  year={2026},
  booktitle={IROS},
}

@inproceedings{luo2024omnigrasp,
  title={Omnigrasp: Grasping Diverse Objects with Simulated Humanoids},
  author={Zhengyi Luo and Jinkun Cao and Sammy Joe Christen and Alexander Winkler and Kris Kitani and Weipeng Xu},
  booktitle={NeurIPS},
  year={2024},
}

@inproceedings{chen2025bodex,
  title={Bodex: Scalable and efficient robotic dexterous grasp synthesis using bilevel optimization},
  author={Chen, Jiayi and Ke, Yubin and Wang, He},
  booktitle={ICRA},
  year={2025},
}

@inproceedings{lum2025crossing,
  title={Crossing the human-robot embodiment gap with sim-to-real rl using one human demonstration},
  author={Lum, Tyler Ga Wei and Lee, Olivia Y and Liu, C Karen and Bohg, Jeannette},
  booktitle={CoRL},
  year={2025}
}

@inproceedings{zhang2025robustdexgrasp,
  title={Robustdexgrasp: Robust dexterous grasping of general objects},
  author={Zhang, Hui and Wu, Zijian and Huang, Linyi and Christen, Sammy and Song, Jie},
  booktitle={CoRL},
  year={2025}
}

@inproceedings{wei2024dro,
  title={$\mathcal{D}(\mathcal{R}, \mathcal{O})$ Grasp: A Unified Representation of Robot and Object Interaction for Cross-Embodiment Dexterous Grasping},
  author={Wei, Zhenyu and Xu, Zhixuan and Guo, Jingxiang and Hou, Yiwen and Gao, Chongkai and Cai, Zhehao and Luo, Jiayu and Shao, Lin},
  booktitle={ICRA},
  year={2025}
}

@inproceedings{lum2024dextrah,
  title={Dextrah-g: Pixels-to-action dexterous arm-hand grasping with geometric fabrics},
  author={Lum, Tyler Ga Wei and Matak, Martin and Makoviychuk, Viktor and Handa, Ankur and Allshire, Arthur and Hermans, Tucker and Ratliff, Nathan D and Van Wyk, Karl},
  booktitle={CoRL},
  year={2024}
}

@inproceedings{wen2024foundationpose,
  title={Foundationpose: Unified 6d pose estimation and tracking of novel objects},
  author={Wen, Bowen and Yang, Wei and Kautz, Jan and Birchfield, Stan},
  booktitle={CVPR},
  year={2024}
}

@inproceedings{ravi2024sam2,
  title={SAM 2: Segment Anything in Images and Videos},
  author={Ravi, Nikhila and Gabeur, Valentin and Hu, Yuan-Ting and Hu, Ronghang and Ryali, Chaitanya and Ma, Tengyu and Khedr, Haitham and R{\"a}dle, Roman and Rolland, Chloe and Gustafson, Laura and Mintun, Eric and Pan, Junting and Alwala, Kalyan Vasudev and Carion, Nicolas and Wu, Chao-Yuan and Girshick, Ross and Doll{\'a}r, Piotr and Feichtenhofer, Christoph},
  booktitle={ICLR},
  year={2025}
}

@inproceedings{huang2023voxposer,
  title={VoxPoser: Composable 3D Value Maps for Robotic Manipulation with Language Models},
  author={Huang, Wenlong and Wang, Chen and Zhang, Ruohan and Li, Yunzhu and Wu, Jiajun and Fei-Fei, Li},
  booktitle={CoRL},
  year={2023}
}

@inproceedings{huang2024rekep,
  title={ReKep: Spatio-Temporal Reasoning of Relational Keypoint Constraints for Robotic Manipulation},
  author={Huang, Wenlong and Wang, Chen and Li, Yunzhu and Zhang, Ruohan and Fei-Fei, Li},
  booktitle={CoRL},
  year={2024}
}

@inproceedings{liang2023code,
  title={Code as Policies: Language Model Programs for Embodied Control},
  author={Liang, Jacky and Huang, Wenlong and Xia, Fei and Xu, Peng and Hausman, Karol and Ichter, Brian and Florence, Pete and Zeng, Andy},
  booktitle={ICRA},
  year={2023}
}

@article{banerjee2024introducing,
  title={Introducing hot3d: An egocentric dataset for 3d hand and object tracking},
  author={Banerjee, Prithviraj and Shkodrani, Sindi and Moulon, Pierre and Hampali, Shreyas and Zhang, Fan and Fountain, Jade and Miller, Edward and Basol, Selen and Newcombe, Richard and Wang, Robert and {Jakob Julian Engel} and {Tomas Hodan}},
  journal={arXiv:2406.09598},
  year={2024}
}

@inproceedings{yang2022oakink,
  title={Oakink: A large-scale knowledge repository for understanding hand-object interaction},
  author={Yang, Lixin and Li, Kailin and Zhan, Xinyu and Wu, Fei and Xu, Anran and Liu, Liu and Lu, Cewu},
  booktitle={CVPR},
  year={2022}
}

@article{andrychowicz2020learning,
  title={Learning dexterous in-hand manipulation},
  author={OpenAI and Andrychowicz, Marcin and Baker, Bowen and Chociej, Maciek and Józefowicz, Rafał and McGrew, Bob and Pachocki, Jakub and Petron, Arthur and Plappert, Matthias and Powell, Glenn and Ray, Alex and Schneider, Jonas and Sidor, Szymon and Tobin, Josh and Welinder, Peter and Weng, Lilian and Zaremba, Wojciech},
  journal={IJRR},
  year={2020},
}

@misc{openai2026gpt55,
  title={GPT-5.5 System Card},
  author={{OpenAI}},
  year={2026},
  howpublished={\url{https://openai.com/index/gpt-5-5-system-card/}},
}

@misc{deepmind2026gemini31pro,
  title={Gemini 3.1 Pro Model Card},
  author={{Google DeepMind}},
  year={2026},
  howpublished={\url{https://deepmind.google/models/model-cards/gemini-3-1-pro/}},
}

@inproceedings{li2025maniptrans,
  title={Maniptrans: Efficient dexterous bimanual manipulation transfer via residual learning},
  author={Li, Kailin and Li, Puhao and Liu, Tengyu and Li, Yuyang and Huang, Siyuan},
  booktitle={CVPR},
  year={2025}
}

@article{mandi2025dexmachina,
  title={Dexmachina: Functional retargeting for bimanual dexterous manipulation},
  author={Mandi, Zhao and Hou, Yifan and Fox, Dieter and Narang, Yashraj and Mandlekar, Ajay and Song, Shuran},
  journal={arXiv:2505.24853},
  year={2025}
}

@inproceedings{kerr2024robot,
  title={Robot see robot do: Imitating articulated object manipulation with monocular 4d reconstruction},
  author={Kerr, Justin and Kim, Chung Min and Wu, Mingxuan and Yi, Brent and Wang, Qianqian and Goldberg, Ken and Kanazawa, Angjoo},
  booktitle={CoRL},
  year={2024}
}

@inproceedings{de2026scaffolding,
  title={Scaffolding dexterous manipulation with vision-language models},
  author={de Bakker, Vincent and Hejna, Joey and Lum, Tyler and Celik, Onur and Taranovic, Aleksandar and Blessing, Denis and Neumann, Gerhard and Bohg, Jeannette and Sadigh, Dorsa},
  booktitle={NeurIPS},
  year={2026}
}

@inproceedings{liu2025dextrack,
  title={Dextrack: Towards generalizable neural tracking control for dexterous manipulation from human references},
  author={Liu, Xueyi and Adalibieke, Jianibieke and Han, Qianwei and Qin, Yuzhe and Yi, Li},
  booktitle={ICLR},
  year={2025}
}

@inproceedings{bharadhwaj2024track2act,
  title={Track2act: Predicting point tracks from internet videos enables generalizable robot manipulation},
  author={Bharadhwaj, Homanga and Mottaghi, Roozbeh and Gupta, Abhinav and Tulsiani, Shubham},
  booktitle={ECCV},
  year={2024},
}

@inproceedings{xu2024flow,
  title={Flow as the cross-domain manipulation interface},
  author={Xu, Mengda and Xu, Zhenjia and Xu, Yinghao and Chi, Cheng and Wetzstein, Gordon and Veloso, Manuela and Song, Shuran},
  booktitle={CoRL},
  year={2024}
}

@inproceedings{wen2023any,
  title={Any-point trajectory modeling for policy learning},
  author={Wen, Chuan and Lin, Xingyu and So, John and Chen, Kai and Dou, Qi and Gao, Yang and Abbeel, Pieter},
  booktitle={RSS},
  year={2024}
}

@article{zhu2026vision,
  title={Vision-based manipulation from single human video with open-world object graphs},
  author={Zhu, Yifeng and Lim, Arisrei and Stone, Peter and Zhu, Yuke},
  journal={Autonomous Robots},
  year={2026},
}

@inproceedings{heppert2024ditto,
  title={Ditto: Demonstration imitation by trajectory transformation},
  author={Heppert, Nick and Argus, Max and Welschehold, Tim and Brox, Thomas and Valada, Abhinav},
  booktitle={IROS},
  year={2024},
}

@inproceedings{li2024okami,
  title={Okami: Teaching humanoid robots manipulation skills through single video imitation},
  author={Li, Jinhan and Zhu, Yifeng and Xie, Yuqi and Jiang, Zhenyu and Seo, Mingyo and Pavlakos, Georgios and Zhu, Yuke},
  booktitle={CoRL},
  year={2024}
}

@inproceedings{wang2022dexgraspnet,
  title={Dexgraspnet: A large-scale robotic dexterous grasp dataset for general objects based on simulation},
  author={Wang, Ruicheng and Zhang, Jialiang and Chen, Jiayi and Xu, Yinzhen and Li, Puhao and Liu, Tengyu and Wang, He},
  booktitle={ICRA},
  year={2023}
}

@inproceedings{chen2024springgrasp,
  title={Springgrasp: Synthesizing compliant, dexterous grasps under shape uncertainty},
  author={Chen, Sirui and Bohg, Jeannette and Liu, C Karen},
  booktitle={RSS},
  year={2024}
}

@inproceedings{lum2024get,
  title={Get a grip: Multi-finger grasp evaluation at scale enables robust sim-to-real transfer},
  author={Lum, Tyler Ga Wei and Li, Albert H and Culbertson, Preston and Srinivasan, Krishnan and Ames, Aaron D and Schwager, Mac and Bohg, Jeannette},
  booktitle={CoRL},
  year={2024}
}

@article{fang2025anydexgrasp,
  title={AnyDexGrasp: General Dexterous Grasping for Different Hands with Human-level Learning Efficiency},
  author={Fang, Hao-Shu and Yan, Hengxu and Tang, Zhenyu and Fang, Hongjie and Wang, Chenxi and Lu, Cewu},
  journal={arXiv:2502.16420},
  year={2025}
}

@inproceedings{yuan2025cross,
  title={Cross-embodiment dexterous grasping with reinforcement learning},
  author={Yuan, Haoqi and Zhou, Bohan and Fu, Yuhui and Lu, Zongqing},
  booktitle={ICLR},
  year={2025}
}

@inproceedings{wei2026one,
  title={One Hand to Rule Them All: Canonical Representations for Unified Dexterous Manipulation},
  author={Wei, Zhenyu and Yao, Yunchao and Ding, Mingyu},
  booktitle={RSS},
  year={2026}
}

@article{wu2026unimorphgrasp,
  title={UniMorphGrasp: Diffusion Model with Morphology-Awareness for Cross-Embodiment Dexterous Grasp Generation},
  author={Wu, Zhiyuan and Zhang, Xiangyu and Chen, Zhuo and Deng, Jiankang and Potamias, Rolandos Alexandros and Luo, Shan},
  journal={arXiv preprint arXiv:2602.00915},
  year={2026}
}

@inproceedings{kedia2026simtoolreal,
  title={SimToolReal: An Object-Centric Policy for Zero-Shot Dexterous Tool Manipulation},
  author={Kedia, Kushal and Lum, Tyler Ga Wei and Bohg, Jeannette and Liu, C Karen},
  booktitle={RSS},
  year={2026}
}

@inproceedings{kuang2026dex4d,
  title={Dex4D: Task-Agnostic Point Track Policy for Sim-to-Real Dexterous Manipulation},
  author={Kuang, Yuxuan and Park, Sungjae and Fragkiadaki, Katerina and Tulsiani, Shubham},
  booktitle={RSS},
  year={2026}
}

@inproceedings{sundaralingam2023curobo,
  title={Curobo: Parallelized collision-free robot motion generation},
  author={Sundaralingam, Balakumar and Hari, Siva Kumar Sastry and Fishman, Adam and Garrett, Caelan and Van Wyk, Karl and Blukis, Valts and Millane, Alexander and Oleynikova, Helen and Handa, Ankur and Ramos, Fabio and Ratliff, Nathan and Fox, Dieter},
  booktitle={ICRA},
  year={2023},
}

@article{mittal2025isaac,
  title={Isaac lab: A gpu-accelerated simulation framework for multi-modal robot learning},
    author={Mittal, Mayank and Roth, Pascal and Tigue, James and Richard, Antoine and Zhang, Octi and Du, Peter and Serrano-Mu{\~n}oz, Antonio and Yao, Xinjie and Zurbr{\"u}gg, Ren{\'e} and Rudin, Nikita and Wawrzyniak, Lukasz and Rakhsha, Milad and Denzler, Alain and Heiden, Eric and Borovicka, Ales and Ahmed, Ossama and Akinola, Iretiayo and Anwar, Abrar and Carlson, Mark T. and Feng, Ji Yuan and Garg, Animesh and Gasoto, Renato and Gulich, Lionel and Guo, Yijie and Gussert, M. and Hansen, Alex and Kulkarni, Mihir and Li, Chenran and Liu, Wei and Makoviychuk, Viktor and Malczyk, Grzegorz and Mazhar, Hammad and Moghani, Masoud and Murali, Adithyavairavan and Noseworthy, Michael and Poddubny, Alexander and Ratliff, Nathan and Rehberg, Welf and Schwarke, Clemens and Singh, Ritvik and Smith, James Latham and Tang, Bingjie and Thaker, Ruchik and Trepte, Matthew and Van Wyk, Karl and Yu, Fangzhou and Millane, Alex and Ramasamy, Vikram and Steiner, Remo and Subramanian, Sangeeta and Volk, Clemens and Chen, CY and Jawale, Neel and Kuruttukulam, Ashwin Varghese and Lin, Michael A. and Mandlekar, Ajay and Patzwaldt, Karsten and Welsh, John and Zhao, Huihua and Anes, Fatima and Lafleche, Jean-Francois and Mo{\"e}nne-Loccoz, Nicolas and Park, Soowan and Stepinski, Rob and Van Gelder, Dirk and Amevor, Chris and Carius, Jan and Chang, Jumyung and Chen, Anka He and Ciechomski, Pablo de Heras and Daviet, Gilles and Mohajerani, Mohammad and von Muralt, Julia and Reutskyy, Viktor and Sauter, Michael and Schirm, Simon and Shi, Eric L. and Terdiman, Pierre and Vilella, Kenny and Widmer, Tobias and Yeoman, Gordon and Chen, Tiffany and Grizan, Sergey and Li, Cathy and Li, Lotus and Smith, Connor and Wiltz, Rafael and Alexis, Kostas and Chang, Yan and Chu, David and Fan, Linxi Jim and Farshidian, Farbod and Handa, Ankur and Huang, Spencer and Hutter, Marco and Narang, Yashraj and Pouya, Soha and Sheng, Shiwei and Zhu, Yuke and Macklin, Miles and Moravanszky, Adam and Reist, Philipp and Guo, Yunrong and Hoeller, David and State, Gavriel},
  journal={arXiv:2511.04831},
  year={2025}
}

@article{schulman2017proximal,
  title={Proximal policy optimization algorithms},
  author={Schulman, John and Wolski, Filip and Dhariwal, Prafulla and Radford, Alec and Klimov, Oleg},
  journal={arXiv:1707.06347},
  year={2017}
}

@article{calli2017yale,
  title={Yale-CMU-Berkeley dataset for robotic manipulation research},
  author={Calli, Berk and Singh, Arjun and Bruce, James and Walsman, Aaron and Konolige, Kurt and Srinivasa, Siddhartha and Abbeel, Pieter and Dollar, Aaron M},
  journal={IJRR},
  year={2017},
}

@inproceedings{ke2025video,
  title={Video depth without video models},
  author={Ke, Bingxin and Narnhofer, Dominik and Huang, Shengyu and Ke, Lei and Peters, Torben and Fragkiadaki, Katerina and Obukhov, Anton and Schindler, Konrad},
  booktitle={CVPR},
  year={2025}
}

@article{romero2022embodied,
  title={Embodied hands: Modeling and capturing hands and bodies together},
  author={Romero, Javier and Tzionas, Dimitrios and Black, Michael J},
  journal={arXiv preprint arXiv:2201.02610},
  year={2022}
}

@inproceedings{chao2021dexycb,
  title={Dexycb: A benchmark for capturing hand grasping of objects},
  author={Chao, Yu-Wei and Yang, Wei and Xiang, Yu and Molchanov, Pavlo and Handa, Ankur and Tremblay, Jonathan and Narang, Yashraj S and Van Wyk, Karl and Iqbal, Umar and Birchfield, Stan and Kautz, Jan and Fox, Dieter},
  booktitle={CVPR},
  year={2021},
}

\clearpage
\clearpage
\appendix

\section{Reinforcement Learning Details}

\subsection{Reward Function Details}
The per-step reward is
\begin{align}
r \;=\;& w_d\,r_{\mathrm{dist}} 
       + w_c\,r_{\mathrm{contact}} + w_l\,r_{\mathrm{lift}} + w_o^{p}\,r^{p}_{\mathrm{obj}} + w_o^{r}\,r^{r}_{\mathrm{obj}},
\end{align}
Reward weights are listed in Table~\ref{tab:rew-weights}.

\paragraphc{Distance reward $r_{\text{dist}}$.} A negative weighted sum of link-to-surface distances, each
clipped at $0.1$\,m:
\[
r_{\mathrm{dist}} = -\sum_{l} \omega_l\,\min\!\big(\lVert \mathbf{a}_l\rVert,\,0.1\big),
\]
with unnormalized per-link weights $\omega_{\text{thumb-tip}} = 8$, $\omega_{\text{other fingertips}} = 4$, $\omega_{\text{palm}} = 0$, and $\omega_l = 1$ for all remaining links. In cross-embodiment training, the weights are masked to active links and normalized to sum to $1$ per environment. For each canonical link $l$ at world position $\mathbf{p}_l$, $\mathbf{a}_l$ is the displacement to the nearest of $M{=}512$ sampled object-surface points, and $\lVert\mathbf{a}_l\rVert$ is its Euclidean norm. The point cloud is stored in the object frame and transformed to the world frame using the tracked object pose.

\paragraphc{Contact reward $r_{\text{contact}}$.} Encourages both thumb contact and an opposing grasp, in which the thumb and at least one non-thumb finger simultaneously contact the object:
\[
r_{\mathrm{contact}} = \mathbf{1}[\text{thumb in contact}] + \mathbf{1}[\text{is\_grasping}],
\]
where a finger is considered to be in contact if its contact force with the object exceeds $\epsilon_F=0.05$\,N, and $\text{is\_grasping}=(\text{thumb in contact})\wedge(\exists\,\text{non-thumb finger in contact})$.

\paragraphc{Lifting reward $r_{\text{lift}}$.} A grasp-gated, scaled reward proportional to the object's height above its post-reset position:
\[
r_{\mathrm{lift}} = \mathbf{1}[\text{is\_grasping}]\cdot \mathrm{clip}\!\big(100\,\Delta h,\ 0,\ 100\,h_{\mathrm{thr}}\big),
\quad \Delta h=\max(0,\ z-z_0),
\]
where $z$ is the object's current height, $z_0$ is its height immediately after reset, and $h_{\mathrm{thr}}=0.05$\,m (so the clip ceiling is $5.0$). The indicator $\mathbf{1}[\text{is\_grasping}]$ ensures that lifting is rewarded only when the object is grasped. Clipping prevents additional reward from throwing the object upward.

\paragraphc{Object stability reward $r_{\text{obj}}$.} Exponential ``stay-put'' bonuses on planar displacement and rotation
from the post-reset pose:
\[
r^{p}_{\mathrm{obj}}=e^{-10\,\lVert \Delta \mathbf{xy}\rVert},
\qquad
r^{r}_{\mathrm{obj}}=e^{-\Delta\theta(\mathbf{q},\mathbf{q}_0)} ,
\]
where $\Delta \mathbf{xy} \in \mathbb{R}^2$ is the planar displacement from the object's post-reset position, and $\Delta \theta(\mathbf{q}, \mathbf{q}_0)=2\arccos(|\langle \mathbf{q},\mathbf{q}_0\rangle|)$ is the geodesic angle between its current and post-reset orientations, represented by unit quaternions. These rewards discourage sliding or rotating the object during contact acquisition, which we found important for stable real-world execution.

\paragraphc{Curriculum / annealing.}
The object-stability weights are linearly annealed from an initial value
to the final value over $T_{\mathrm{cur}}$ per-environment control steps:
$w(t)=w_{\mathrm{init}}(1-\rho)+w_{\mathrm{final}}\rho$,
$\rho=\min(1,t/T_{\mathrm{cur}})$. We anneal $w_o^{p},w_o^{r}$ from
$w_{\text{init}}=2.0$ to $w_{\text{final}}=10.0$ over $T_{\mathrm{cur}}=10^5$ steps. The distance, contact, and lifting weights are not annealed.

\begin{table}[h]
\centering
\caption{Reward weights.}
\label{tab:rew-weights}
\begin{tabular}{llc}
\toprule
Symbol & Term & Value \\
\midrule
$w_d$ & distance & $16.0$ \\
$w_c$ & contact & $4.0$ \\
$w_o^{p}$ & object position stability & $10.0$ (init.\ $2.0$) \\
$w_o^{r}$ & object rotation stability & $10.0$ (init.\ $2.0$) \\
$w_l$ & lifting & $1.0$ \\
$T_{\text{cur}}$ & annealing steps & $10^5$ \\
\bottomrule
\end{tabular}
\end{table}

\subsection{Cross-Embodiment Unified Policy Representation}
\label{sec:rl-canonical}

\paragraphc{Canonical joint space.}
We map every hand onto a fixed \emph{canonical} joint layout consisting of $J_{\mathrm{can}}=26$ slots partitioned into five contiguous per-finger blocks (6 for the thumb and 5 for each other finger). The wrist is excluded from this layout and controlled separately through a shared 6-DoF floating-base interface. Each hand fills only the slots corresponding to its native joints, and the remaining slots are masked. The number of active slots equals the hand's DoF: XHand $12$, Allegro $16$, and Sharpa Wave $22$ (Table~\ref{tab:canonical-joints}).

\begin{table*}[h]
\centering
\caption{\textbf{Canonical 26-DoF joint layout and per-hand instantiation.} Each finger is allocated a fixed number of slots in the canonical layout (6 for the thumb, 5 for each non-thumb finger); slots are filled by a hand's native joint when present, and masked (\texttt{[]}) otherwise. F, A, and R denote flexion, abduction/adduction, and axial rotation respectively. Reading order within each finger is base $\to$ tip.}

\label{tab:canonical-joints}
\begin{tabular}{lccccccc}
\toprule
Hand & DoF & Thumb (6) & Index (5) & Middle (5) & Ring (5) & Pinky (5) \\
\midrule
Allegro     & 16 & \texttt{RAF[\,]F[\,]} & \texttt{R[\,]FFF} & \texttt{R[\,]FFF} & \texttt{R[\,]FFF} & \texttt{[\,][\,][\,][\,][\,]} \\
Sharpa Wave & 22 & \texttt{[\,]AFAFF}    & \texttt{[\,]AFFF} & \texttt{[\,]AFFF} & \texttt{[\,]AFFF} & \texttt{FAFFF} \\
XHand      & 12 & \texttt{[\,]AF[\,]F[\,]} & \texttt{[\,]AFF[\,]} & \texttt{[\,][\,]FF[\,]} & \texttt{[\,][\,]FF[\,]} & \texttt{[\,][\,]FF[\,]} \\
\bottomrule
\end{tabular}
\end{table*}

\paragraphc{Canonical link space.}
For the canonical link ordering, we map each robot's native link positions (obtained via forward kinematics from its URDF) onto a fixed $L_{\mathrm{can}}=21$ keypoint skeleton following the standard hand topology used in MANO~\cite{romero2022embodied}: one palm keypoint (canonical index $0$) plus four keypoints per finger ($5 \times 4$) ordered base-to-tip, with fingertips at canonical indices $\{4, 8, 12, 16, 20\}$. Link observations and link-based rewards use this canonical ordering.

\paragraphc{Per-robot canonical mapping.}
Each robot $r$ stores an index map $\pi_r$ and a sign vector
$s_r$ with entries in $\{-1,+1\}$ that translate between its native joints and the
canonical slots. The policy reads and writes in the \emph{canonical} space.
At actuation, the canonical hand command $a^{\mathrm{hand}}\!\in\!
\mathbb{R}^{J_{\mathrm{can}}}$ is gathered into native joint order and
sign-corrected to account for differences in native joint-axis conventions:
\[
a^{\mathrm{native}}_r \;=\; a^{\mathrm{hand}}[\pi_r]\odot s_r .
\]

\paragraphc{Masks and padding.}
For each robot we form binary masks
$\delta^{\mathrm{joint}}\!\in\!\{0,1\}^{26}$ and
$\delta^{\mathrm{link}}\!\in\!\{0,1\}^{21}$ that flag occupied canonical
slots. Both masks are appended to the observation so that the policy can identify active dimensions. Unoccupied joint slots are filled with zeros, and missing link positions are filled using the nearest existing link.

\subsection{Observations}
The observation concatenates three blocks: a 5-frame proprioceptive history, single-frame hand-object displacement vectors, and joint and link masks. The policy does not receive the object pose explicitly. Object information enters only through the per-link nearest-surface displacement vectors $\mathbf{a}_l$ defined in the distance reward above. Within the history buffer, frames are ordered from oldest to newest, $[o_{t-K+1}, \ldots, o_t]$ with history length $K = 5$. At episode start, the buffer is padded with copies of the first observed frame. The full observation breakdown is given in Table~\ref{tab:obs}.

\begin{table}[h]
\centering
\caption{Observation components. $K\!=\!5$. }
\label{tab:obs}
\begin{tabular}{lll}
\toprule
Component & Composition & Dim \\
\midrule
Joint positions      & $26$ canonical $\times\,K$ frames        & $26K$ \\
Joint targets   & $26$ canonical $\times\,K$ frames        & $26K$ \\
Link positions       & $21$ links $\times\,3\times\,K$ frames    & $63K$ \\
\midrule
Hand-object vectors   & $21$ links $\times\,1$ nearest pt $\times\,3$ & $63$ \\
\midrule
Joint mask           & active canonical joints                  & $26$ \\
Link mask            & active canonical links                   & $21$ \\
\bottomrule
\end{tabular}
\end{table}

\paragraphc{Joint positions and joint targets.} Joint positions and previous-step joint targets are reordered into the canonical 26-joint layout and padded with zeros for joints not present on the active hand.

\paragraphc{Link positions.} We use forward kinematics to compute the 3D world positions of links in the canonical 21-keypoint topology, discarding their orientations. Missing keypoints are padded using the nearest existing link, as described in Section~\ref{sec:rl-canonical}.

\paragraphc{Hand-object vectors (following~\cite{zhang2025robustdexgrasp}).}
For each of the $21$ canonical links we compute the vector to its
\emph{nearest} point on the object surface point cloud
($N\!=\!512$ points sampled per object), giving a $21\times3$ tensor.

\paragraphc{Normalization.}
There is no in-environment normalization. \texttt{rl\_games} normalizes observations and value targets using running means and standard deviations. Observations are clipped to $[-5,5]$.

\subsection{Policy Action}
\paragraphc{Action space.}
The action is
$a=[\,a^{\mathrm{EE}}(6),\ a^{\mathrm{hand}}(J_{\mathrm{can}})\,]$, i.e.
$6+26=32$ dimensions.
The first $6$ dimensions specify an end-effector \emph{delta} (3 translation components and 3 Euler angles) for the shared floating base, converted to
the $6$ virtual base-joint targets.
The hand block is in canonical space and remapped per robot as in Section~\ref{sec:rl-canonical}.

\paragraphc{Range, scaling, and clipping.}
Raw policy actions are clipped to $[-1, 1]$ before being rescaled into physical commands. All controllers operate in residual (delta-target) mode: each scaled delta is added to the previous control target, which is then clamped to the corresponding physical limit. For the hand, per-joint deltas are rescaled using each joint's upper and lower limits so that an action of $\pm 1$ corresponds to a full-range increment. The end-effector deltas are scaled to $\pm0.008$\,m per step for translation and $\pm0.052$\,rad per step for rotation, then composed with the previous target pose.

\paragraphc{Control frequency.}
Simulation uses a timestep of $\mathrm{dt}=1/120$\,s. With \texttt{decimation}$=12$, each action is held for $12$ physics substeps, giving a control rate of $10$\,Hz.
\subsection{Reset and Initialization}

\paragraphc{Length and termination.} We use a fixed episode horizon of $H = 60$ control steps (corresponding to $6.0$\,s at $10$\,Hz). An episode terminates when the step counter reaches $H$. At the start of training, episode step counters across the 4{,}096 parallel environments are initialized uniformly at random in $\{0, 1, \dots, H{-}1\}$ rather than synchronously at zero. This staggers episode resets so that rollout batches contain transitions from different phases (approach, contact, and lifting), improving sample diversity and stabilizing early training.

\paragraphc{Reset distribution.}
\label{sec:rl-episode}
At scene initialization, each environment randomly selects a robot embodiment and an object. At each reset, it samples a valid pair (object resting pose, initial hand configuration) from the per-object grasp dataset. We further augment the initial state with a random planar translation $\Delta \mathbf{p}_{xy} \sim \mathcal{U}([-0.05, 0.05]^2)$\,m and a random yaw $\phi \sim \mathcal{U}(-\pi, \pi)$ about the world $z$-axis. The same transformation is applied to both components, preserving their relative geometry. 
\subsection{Domain Randomization}
We apply domain randomization to encourage robustness and sim-to-real transfer. All physics parameters are resampled at every environment reset. Friction and restitution are set directly to values sampled from the listed ranges. Mass, scale, and actuator gains are scaled relative to their nominal values for each robot or object. The nominal object density is $\rho=200\,$kg/m$^3$, which determines each object's baseline mass before the $\times[0.5, 3.0]$ mass scale is applied.

\begin{table*}[h]
\centering
\caption{\textbf{Domain randomization ranges.} All distributions are uniform. Multiplicative ranges ($\times$) scale the per-object or per-robot nominal value; absolute ranges directly set the parameter value.}
\label{tab:dr}
\setlength{\tabcolsep}{8pt}
\renewcommand{\arraystretch}{1.15}
\begin{tabular}{llll}
\toprule
Group & Parameter & Range & Type \\
\midrule
\multirow{4}{*}{Object}
 & Scale                   & $[0.96, 1.04]$ & $\times$ nominal \\
 & Mass                    & $[0.5, 3.0]$   & $\times$ nominal (from $\rho \cdot V \cdot \text{scale}^3$) \\
 & Static/dynamic friction & $[0.5, 1.5]$   & absolute \\
 & Restitution             & $[0.0, 0.1]$   & absolute \\
\midrule
\multirow{2}{*}{Robot links}
 & Static/dynamic friction & $[0.5, 2.0]$   & absolute \\
 & Restitution             & $[0.1, 0.5]$   & absolute \\
\midrule
\multirow{2}{*}{Actuators}
 & Stiffness ($k_p$)       & $[0.5, 2.0]$   & $\times$ nominal \\
 & Damping ($k_d$)         & $[0.3, 3.0]$   & $\times$ nominal \\
\midrule
\multirow{2}{*}{Perturbation}
 & Push force (max)        & $2.0$\,N       & sampled uniformly per axis \\
 & Application offset      & $\pm 0.05$\,m  & from object center \\
\bottomrule
\end{tabular}
\end{table*}

\paragraphc{Nominal actuator gains.}
The floating arm base has $k_p = [1000, 1000, 1000, 300, 300, 300]$ and $k_d = [100, 100, 100, 15, 15, 15]$ for its six virtual joints. The per-hand $k_p$ and $k_d$ values are obtained through system identification on the physical hardware.

\subsection{Contact Prediction Details}\label{sec:appendix-contact-details}
We train a separate contact predictor using rollouts from the frozen policy after convergence. The predictor is a 2-layer MLP with a hidden size of 256 and ELU activation. It takes joint positions, joint targets, and link positions over the most recent $k=5$ steps and outputs five per-finger contact logits. We train it with binary cross-entropy. The ground-truth label for each finger is 1 if the contact-sensor force exceeds $\epsilon_F = 0.05$\,N.

\paragraphc{Training.} At each epoch, we collect 120 steps using deterministic actions from the frozen policy across 4{,}096 parallel environments. We train the predictor on the resulting in-memory buffer in shuffled mini-batches of 4{,}096 with Adam at a learning rate of $3\times 10^{-4}$. We train for 200 epochs in total.

\paragraphc{Deployment.} At inference, contact probabilities are obtained by sigmoid activation and thresholded at $0.5$. 
We set \texttt{is\_grasping} to true when the thumb and at least one non-thumb finger are predicted to be in contact.
We switch to tracking when \texttt{is\_grasping} remains true for five consecutive control steps.

\subsection{Additional Training Details}
\paragraphc{Network architecture.} We use a feed-forward MLP with hidden sizes $[512, 512, 256, 128]$. Training hyperparameters are listed in Table~\ref{tab:ppo}.

\label{sec:rl-ppo}

\begin{table}[htb]
\centering
\caption{PPO hyperparameters.}
\label{tab:ppo}
\begin{tabular}{ll}
\toprule
Parameter & Value \\
\midrule
Parallel environments & $4096$ \\
Horizon (rollout) length & $60$\\
Minibatch size & $8192$ \\
Mini-epochs per update & $5$ \\
Discount $\gamma$ & $0.95$ \\
GAE $\lambda$ (\texttt{tau}) & $0.90$ \\
Learning rate & $3\times10^{-4}$, adaptive (KL) \\
KL threshold & $0.016$ \\
PPO clip $\epsilon$ & $0.2$ \\
Value loss coef.\ (\texttt{critic\_coef}) & $4$ \\
Value clipping & enabled \\
Entropy coef. & $0.0$ \\
Bounds-loss coef. & $1\times10^{-4}$ \\
Gradient norm clip & $1.0$ \\
Normalize input / value / advantage & yes / yes / yes \\
Mixed precision & no \\
Max epochs & $2000$ \\
Seed & $42, 84, 126$ \\
\bottomrule
\end{tabular}
\end{table}

\begin{figure}[t]
    \centering
    \includegraphics[width=0.5\columnwidth]{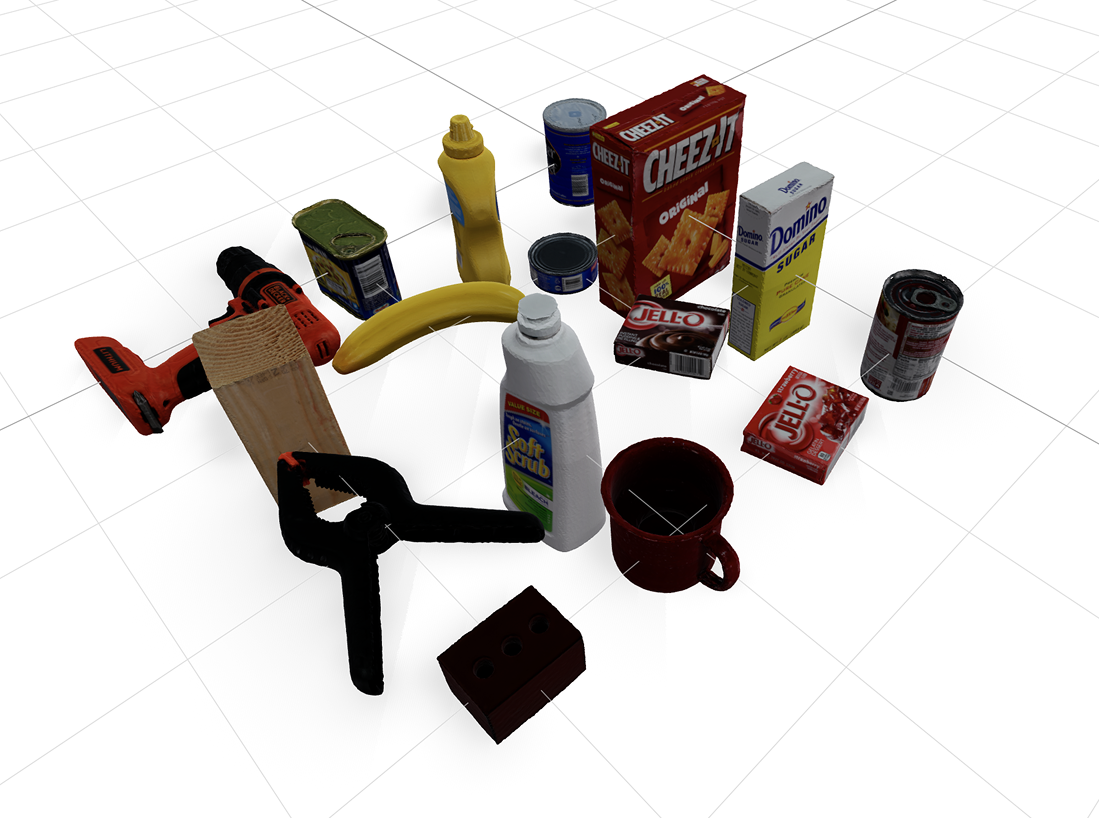}
    \caption{Simulation training assets.}
    \label{fig:sim-assets}
\end{figure}

\paragraphc{Object set.} We use $16$ YCB objects, as illustrated in Figure~\ref{fig:sim-assets}. Each object has $2$--$7$ distinct stable resting poses on the table ($\sim\!100$ object--pose configurations per hand).

\paragraphc{Grasp data generation and filtering.}
For each (object, resting-pose) pair, we generate grasp candidates with BODex~\cite{chen2025bodex}. BODex uses a heuristic seed sampler to propose an initial hand configuration, which we keep for the reset distribution (Section~\ref{sec:rl-episode}). The full bilevel grasp optimization then refines this seed into a converged \emph{reference} grasp. We validate the reference grasps with BODex's official evaluation code and retain 4{,}000 successful (initial configuration, reference grasp) pairs per (object, resting-pose) pair.

\paragraphc{Hardware and training duration.}
We train each policy in IsaacLab on a single NVIDIA L40S GPU for 2000 epochs, taking 24 hours. This corresponds to approximately 491.5 million environment steps, with 60 steps across 4096 parallel environments per epoch.

\section{Future-Aware Planning Details}
\label{sec:vlm-select}
At deployment, the planner identifies a task-relevant contact region and then generates candidate initializations within this region. A vision-language model (VLM) provides the task-aware affordance selection, while candidate generation and trajectory selection are performed geometrically and kinematically.

\begin{figure}[htb]
    \centering
    \includegraphics[width=\linewidth]{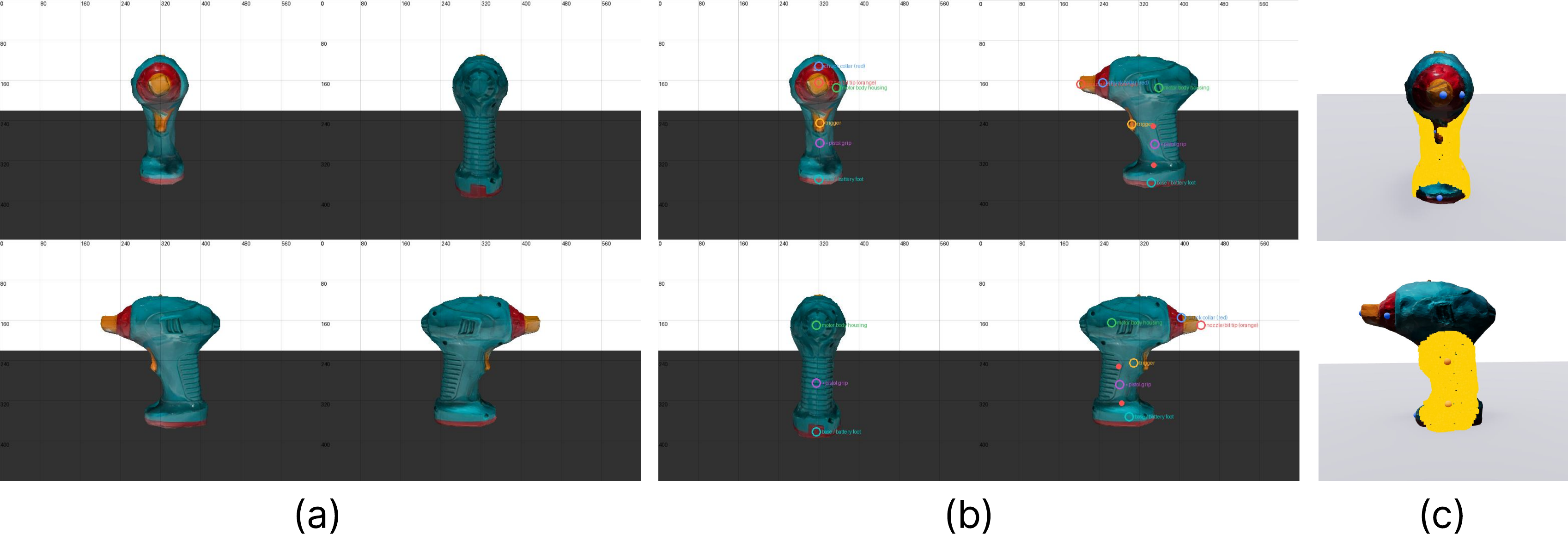}
    \caption{\textbf{Task-aware region selection for \texttt{Drill Use}.} (a)~Four rendered views provided to the VLM. (b)~The VLM identifies functional parts and selects the pistol grip as the task-relevant holding region, with multiple grasp points spanning the grip. (c)~Back-projected grasp points define the contact region (yellow), from which candidate initializations are sampled.}
    \label{fig:vlm-affordance}
\end{figure}

\paragraphc{Task-aware region selection.}
We render the object at its observed pose on the table from four side views (Figure~\ref{fig:vlm-affordance} (a)). The views use textured rendering with a low camera elevation and are overlaid with a light pixel grid for image-coordinate reference. The VLM is prompted to identify the object's functional parts, mark each visible part in each view, and designate the single part to be grasped for the specified task (Figure~\ref{fig:vlm-affordance} (b)). For the designated part, the VLM additionally predicts multiple points spanning its extent. All predicted points are back-projected onto the object mesh by ray casting. For each functional part, its annotations across views are clustered in the object frame, and the mean of the largest cluster is used as the part location. Predicted grasp points whose back-projected locations are inconsistent with the designated part are discarded. We define the task-relevant contact region as the union of $3\,$cm surface neighborhoods around the remaining grasp points (Figure~\ref{fig:vlm-affordance} (c)). 

\paragraphc{Candidate generation.}
Following the initialization stage of BODex~\cite{chen2025bodex}, we sample candidate hand poses and joint configurations from surface points within the selected contact region. We include the table in the collision world and reject candidate initializations that collide with the table or other scene geometry, as well as ill-posed poses. We retain $N=10$ valid candidate initializations by default. These candidates serve as approximate initial configurations for the contact-acquisition primitive rather than final grasps.

\paragraphc{Trajectory expansion and selection.}
For each remaining candidate, we plan a collision-free approach from the robot's initial configuration to the initial grasp pose with cuRobo~\cite{sundaralingam2023curobo}. We then assume a rigid hand-object attachment and compute the hand trajectory along the desired object motion $\tau_o$, solving IK along this trajectory to recover the full arm-hand motion. Candidates whose approach or tracking segment is infeasible are discarded. Among the remaining candidates, we select the trajectory with the smallest arm motion,
\[
c^{\mathrm{ik}}_i \;=\; \sum_{t=1}^{T} \big\lVert\, q^{\mathrm{arm}}_{i,t} - q^{\mathrm{arm}}_{i,t-1} \,\big\rVert_2 ,
\]
where $q^{\mathrm{arm}}_{i,t}$ denotes the arm configuration obtained by IK for candidate $i$ at timestep $t$.
       
\paragraphc{Prompt and example.}
We use Claude Opus 4.7 or Claude Opus 5 via the Anthropic API. The prompt specifies the manipulation task and provides generic guidance distinguishing the holding region from the working end of an object. Given the four rendered views, the model returns a JSON list of labeled object parts, with one pixel coordinate for each visible part in each view and a binary \texttt{grasp} flag indicating the part selected for grasping. Figure~\ref{fig:vlm-affordance} shows an example from a \texttt{Drill Use} trial. The VLM identifies the pistol grip as the task-relevant holding region, and the subsequent candidate generation is restricted to this region.

\begin{tcolorbox}[colback=gray!5, colframe=gray!40, fontupper=\small\ttfamily, breakable, enhanced]
You are labeling the functional KEYPOINTS of an object for a dexterous robot, then choosing which one to grasp.\\
Task: \{task\_name\}\\
\\
The object may have a designated HOLDING REGION (shaped to fit a hand) vs a WORKING END (does the task). Grasp the HOLDING REGION. [...]\\
\\
You are given 4 rendered views of the SAME object standing on a table, from different angles. Each image is 640x480 px with a light coordinate grid [...]\\
\\
Step 1 --- Identify the object's distinct FUNCTIONAL KEYPOINTS (the named parts a person would talk about). Give each a short label.\\
Step 2 --- For EACH keypoint, mark its pixel in EVERY view where it is clearly visible. All picks for one keypoint must be the SAME physical point, so we can triangulate it in 3D.\\
Step 3 --- Set "grasp": true on the ONE keypoint the hand should converge on to hold the object for this task, and false on the rest.\\
Step 4 --- For that grasp keypoint ONLY, also give "extent": 2--4 extra pixel picks that span the full length of the holding part (e.g. both ends and the middle of a grip or handle). Together with the keypoint they define WHERE the fingers may land, so cover the whole holdable part but do NOT extend into the working end or onto other parts.\\
\\
Respond with ONLY a JSON object: \{"keypoints": [\{"label": <str>, "grasp": <bool>, "picks": [\{"view": <int>, "u": <int>, "v": <int>\}, ...], "extent": [\{"view": <int>, "u": <int>, "v": <int>\}, ...]\}, ...], "reason": "<why>"\}\\
("extent" only on the grasp keypoint; omit it elsewhere.)
\end{tcolorbox}

\section{Real-World Grasping Experiment Details}

This section provides per-object real-world grasping results across the three arm-hand platforms (Table~\ref{tab:per-object-success}). Tianji-XHand (48/50, 96\%) achieves higher overall success than both Franka platforms (45/50 each, 90\%). We attribute this to the larger reachable workspace of the Tianji arm relative to our Franka setup.

\begin{table*}[h]
  \centering
  \caption{Per-object grasp success across embodiments (out of 5 trials).
  Objects are abbreviated as O1-O10: O1 \texttt{mustard-bottle}, O2 \texttt{bleach-cleanser}, O3 \texttt{drill-red},
  O4 \texttt{tomato-soup}, O5 \texttt{meat-can}, O6 \texttt{red-bottle}, O7 \texttt{drill-green}, O8 \texttt{apple-tea}, O9 \texttt{eraser}, O10 \texttt{maize}. Objects O1--O5 are seen during training; O6--O10 are unseen.}
  \label{tab:per-object-success}
  \begin{tabular}{lcccccccccc|c}
  \toprule
  Embodiment & O1 & O2 & O3 & O4 & O5 & O6 & O7 & O8 & O9 & O10 & Total (\%) \\
  \midrule
  Franka-Sharpa & 4 & 4 & 5 & 5 & 5 & 5 & 4 & 3 & 5 & 5 & 90 \\
  Franka-XHand  & 5 & 5 & 5 & 4 & 5 & 3 & 5 & 4 & 5 & 4 & 90 \\
  Tianji-XHand  & 5 & 5 & 4 & 4 & 5 & 5 & 5 & 5 & 5 & 5 & 96 \\
  \bottomrule
  \end{tabular}
\end{table*}

\section{Real-World Full-Pipeline Experiment Visualizations}
\begin{figure}[t]
    \centering
    \includegraphics[width=\linewidth]{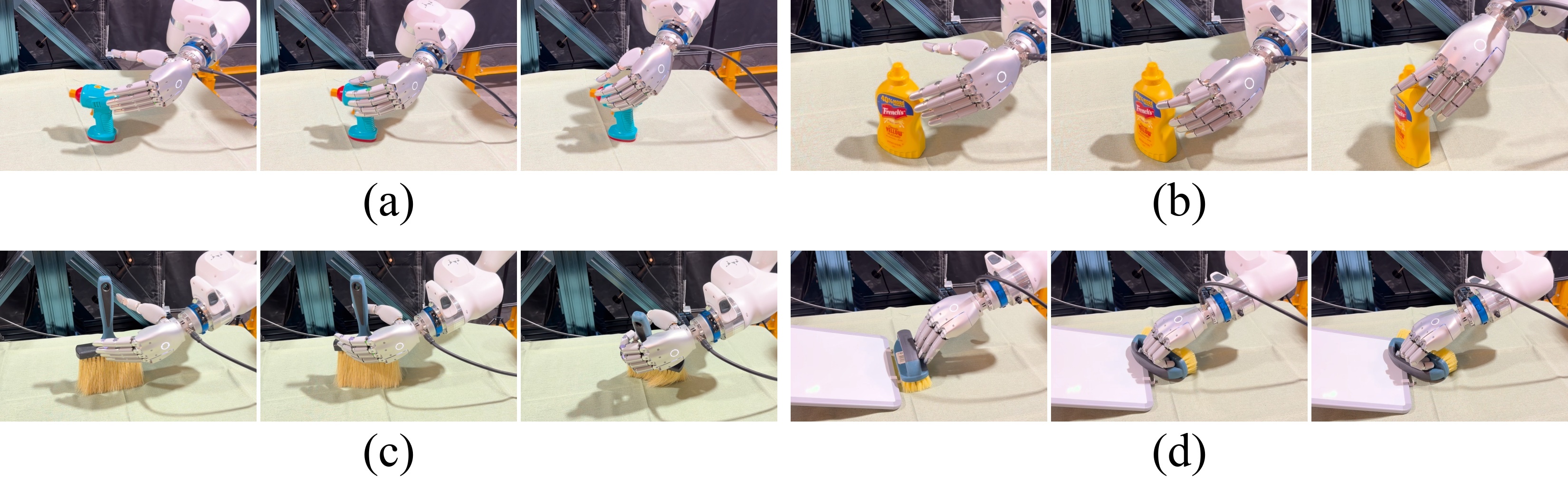}
    \caption{\textbf{Qualitative comparison on real-world tasks.} Subfigures (a)-(b) are from BODex, and (c)-(d) are from DRO. }
    \label{fig:real-tasks}
\end{figure}

We present qualitative comparisons of the full pipeline against the DRO and BODex baselines in Figure~\ref{fig:real-tasks}, illustrating three recurring failure modes: (i) grasping at an unintended contact region that does not support the downstream trajectory (Figure~\ref{fig:real-tasks}(d)), (ii) displacing the object during approach (Figure~\ref{fig:real-tasks}(a, b)), and (iii) forming a cage-like grasp without firm opposing finger-thumb contact (Figure~\ref{fig:real-tasks}(c)). In contrast, \ours addresses these failure modes by (i) adjusting the wrist and fingers locally to recover from imperfect pre-grasp poses produced by upstream planning, (ii) reacting to object displacement during approach, and (iii) establishing firm opposing finger-thumb contact rather than caging the object loosely. We refer the reader to the supplementary video or our project website for video demonstrations of real-world performance.

\section{Simulation Grasping Evaluation Protocol and Baseline Details}
\label{sec:appendix-baseline-protocol}
The grasp success rates we report for BODex~\cite{chen2025bodex} and DRO~\cite{wei2024dro} in Table~\ref{tab:main} of the main paper are lower than those reported in their original publications. This section clarifies the source of this gap and presents a controlled reproduction.

\subsection{Protocol Comparison with BODex}
BODex's original evaluation places each scaled object in a \emph{floating state without a table}. Ten grasps are synthesized per (object, scale, pose) instance. The hand is initialized at the pre-grasp pose $\mathbf{x}_p$ and driven to the squeezed pose $\mathbf{x}_s$. Gravity is then applied along the positive and negative directions of each of the three coordinate axes. A grasp counts as successful only if the object's translation and rotation deviations remain within $5\,$cm and $15^\circ$ for at least $3\,$s under \emph{all six} directions. In contrast, our protocol places objects on a tabletop under gravity from the start of the trial. Both seed-pose initialization (used identically across all methods) and grasp execution must avoid hand-table collisions throughout the trial. This setting is closer to real-world deployment, where objects rest on a workspace surface and the robot must operate within the resulting constraints. In summary, the two protocols differ in two respects:
\begin{itemize}[leftmargin=*, itemsep=1pt, topsep=2pt, partopsep=0pt, parsep=0pt]
    \item \textbf{Generation phase.} BODex synthesizes grasps without considering a table. Our setup adds hand-table collision constraints to both initial-pose sampling and the bilevel optimization.
    \item \textbf{Evaluation phase.} BODex evaluates grasps in a floating-hand state with gravity applied after squeezing along all six directions. Our setup places the object on a table under $-z$ gravity from the start of the trial.
\end{itemize}

\subsection{Reproducing BODex under Its Evaluation Protocol}
To isolate the effect of the generation procedure, we use BODex's officially released code with its default parameters (in the \texttt{main} branch), synthesize 20 initial poses per object under two conditions, and evaluate both under BODex's original protocol:
\begin{itemize}[leftmargin=*, itemsep=1pt, topsep=2pt, partopsep=0pt, parsep=0pt]
\item \textbf{Floating generation} (BODex default): initial-pose sampling and bilevel optimization ignore the table.
\item \textbf{Tabletop generation} (ours): initial-pose sampling and bilevel optimization avoid hand-table collisions.
\end{itemize}

\begin{table}[h]
\centering
\caption{\textbf{Effect of generation-time table constraints on BODex grasp success rate (\%).} Allegro right hand, 4 object scales, evaluated under BODex's original protocol. Only the generation phase differs between the two columns. Objects are taken from DexGraspNet~\cite{wang2022dexgraspnet}, normalised so that each object's bounding-box diagonal measures $1\,$m, and then rescaled to four target diagonals (\texttt{@006}=$0.06\,$m, \texttt{@008}=$0.08\,$m, \texttt{@010}=$0.10\,$m, \texttt{@012}=$0.12\,$m).}
\label{tab:bodex-generation-ablation}
\setlength{\tabcolsep}{8pt}
\begin{tabular}{lccccccc}
\toprule
Scale & \texttt{@006} & \texttt{@008} & \texttt{@010} & \texttt{@012} & avg. \\
\midrule
Floating & 69.79 & 73.79 & 72.09 & 67.26 & 70.73 \\
Tabletop & 53.29 & 60.90 & 60.05 & 54.28 & 57.13 \\
\bottomrule
\end{tabular}
\end{table}

Table~\ref{tab:bodex-generation-ablation} shows that tabletop-aware generation alone, with evaluation held fixed, drops the success rate from $70.7\%$ to $57.1\%$. This degradation is attributable purely to the harder generation problem: the init-pose distribution changes and the bilevel optimization becomes more constrained. The floating-generation result is below the $79.9\%$ reported in~\cite{chen2025bodex} because the released code's \texttt{main} branch differs from the paper's configuration (preserved in the \texttt{baseline} branch) in object mass ($100\,$g vs.\ $30\,$g) and joint stiffness ($k_p{=}5$ vs.\ $k_p{=}2$); rerunning floating evaluation with the paper's configuration (object mass $30\,$g and $k_p{=}2$) yields $77.2\%$, with the residual gap likely due to seed variance. Following the authors' recommendation, Table~\ref{tab:bodex-generation-ablation} and Figure~\ref{fig:bodex-scatter} use the \texttt{main}-branch defaults; this choice does not affect the conclusions above, since both rows of Table~\ref{tab:bodex-generation-ablation} share the same configuration.

\subsection{Evaluation-Phase Comparison}
\begin{figure}[t]
    \centering
    \includegraphics[width=0.6\linewidth]{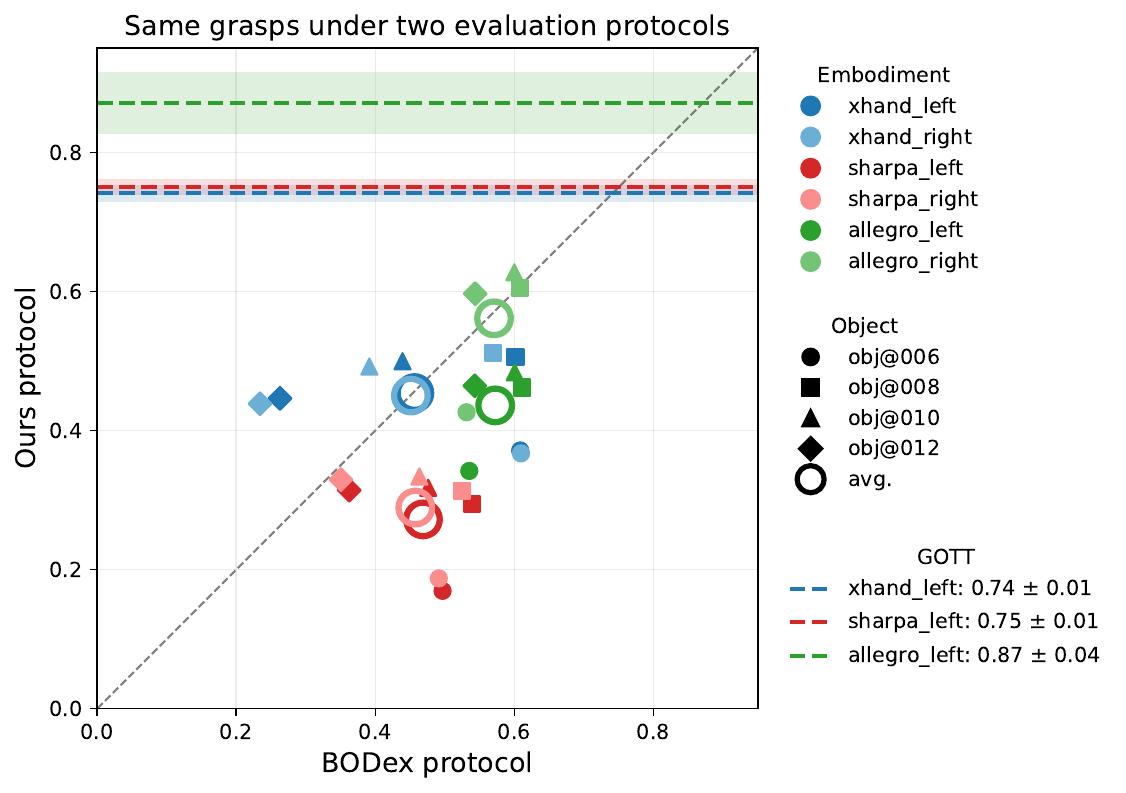}
    \caption{\textbf{Evaluation protocol comparison.} The same BODex grasps evaluated under BODex's original protocol and our tabletop protocol. }
    \label{fig:bodex-scatter}
\end{figure}

We further evaluate the same set of BODex-generated grasps under both protocols. Figure~\ref{fig:bodex-scatter} plots, for each (embodiment, object set) pair, BODex's success rate under its original protocol (x-axis) against the success rate of the \emph{same grasps} under our tabletop protocol (y-axis). The points cluster close to the diagonal, indicating that the two evaluation protocols give similar success-rate estimates. Reference lines mark \ours's tabletop success rates of 0.74 (XHand), 0.75 (Sharpa), and 0.87 (Allegro), which lie well above BODex's results.
Even when BODex grasps are evaluated under its original floating-hand protocol, \ours outperforms BODex on every embodiment, confirming that the gap in Table~\ref{tab:main} of the main paper is not an artifact of the evaluation choice.

\subsection{Protocol Comparison with DRO}
DRO~\cite{wei2024dro} evaluates its grasps in Isaac Gym, where a controller closes the hand around the object. The grasp is then perturbed by sequential forces along the positive and negative directions of each coordinate axis, with a magnitude corresponding to an acceleration of $0.5\,\mathrm{m/s^2}$ (i.e., $F = m_{\text{obj}}(0.5\,\mathrm{m/s^2})$). A grasp is successful if the object's displacement stays below 2\,cm. Our protocol differs in two ways. First, we require the grasp to lift the object against gravity rather than resist a brief perturbation. The gravitational load $F_g = m_{\text{obj}}(9.8\,\mathrm{m/s^2})$ is roughly 20$\times$ the perturbation force $F$ used in DRO's protocol and is sustained throughout the trial. Second, our protocol places the object on a table from the start, whereas DRO's grasp synthesis does not account for table collisions, which limits its performance under our protocol.

\subsection{DRO Baseline Variants}
\label{sec:appendix-dro}

Our main table (Table~\ref{tab:main}) uses DRO's~\cite{wei2024dro} released checkpoint. DRO is trained across embodiments and transfers zero-shot to new hands given their URDFs, but XHand and Sharpa Wave are not part of its training set. For a fair comparison, we additionally fine-tune or retrain DRO on our BODex-generated YCB grasp set. Specifically, we run BODex's tabletop-aware bilevel optimization from the \emph{same} initial hand configurations used by \ours and keep the 4{,}000 grasps that pass BODex's own validity check as the training set. All variants are evaluated under our tabletop protocol (Table~\ref{tab:dro-variants}). All fine-tuning and retraining variants underperform the released checkpoint on average across hands. We attribute this to our smaller 16-object training set with tabletop-constrained grasps, which differs from DRO's original training distribution. We therefore keep the released checkpoint in the main table as the strongest DRO variant overall. Notably, even on the natively supported Allegro hand, DRO reaches 43.2\% versus our 87.1\%, indicating that the gap primarily reflects the benefit of closed-loop contact acquisition over open-loop grasp execution under the same tabletop protocol.

\begin{table}[h]
  \centering
  \caption{\textbf{DRO variants.} Lifting success rate (\%) under our tabletop protocol; best per column in bold.}
  \label{tab:dro-variants}
  \footnotesize
  \setlength{\tabcolsep}{6pt}
  \begin{tabular}{lcccc}
    \toprule
    DRO variant       & Allegro & Sharpa & XHand & Av. \\
    \midrule
    Released ckpt     & \textbf{43.2} & 47.0  & \textbf{31.3} & \textbf{40.5} \\
    Fine-tune 1 ep.   & 21.6 & \textbf{52.2} & 31.1 & 35.1\\
    Fine-tune 5 ep.   & 20.0 & 37.9 & 32.0 & 30.0\\
    Retrain 5 ep.     & 19.8 & 40.6 & 32.6 & 31.0\\
    Retrain 40 ep.    & 18.6 & 35.5 & 31.3 & 28.5\\
    \bottomrule
  \end{tabular}
\end{table}

\section{Additional Experiments}
\label{sec:appendix-extras}

\subsection{Runtime Breakdown}
\label{sec:appendix-runtime}

Table~\ref{tab:runtime} reports the per-stage runtime averaged over 10 rollouts. One-time initialization (4-view rendering, VLM-based affordance reasoning, candidate generation, IK filtering, and cuRobo planning) totals about 29.6\,s. Online execution is real-time: FoundationPose runs at 45\,Hz on an RTX~4090, and the contact-acquisition and contact-prediction networks take 0.752\,ms and 0.085\,ms per step, respectively, supporting the 10\,Hz control loop.

\begin{table}[htb]
  \centering
  \caption{Per-stage runtime, averaged over 10 rollouts.}
  \label{tab:runtime}
  \begin{tabular}{lc}
    \toprule
    Stage & Time \\
    \midrule 
    4-view rendering               & 0.11 $\pm$ 0.01\,s \\
    VLM-based affordance reasoning & 14.9 $\pm$ 0.9\,s \\
    Candidate generation           & 0.76 $\pm$ 0.01\,s \\
    IK filtering                   & 3.95 $\pm$ 0.06\,s \\
    cuRobo planning ($N{=}10$)     & 9.6 $\pm$ 0.5\,s \\
    \midrule
    FoundationPose                 & 45\,Hz \\
    Contact acquisition (per step) & 0.752\,ms \\
    Contact predictor (per step)   & 0.085\,ms \\
    \bottomrule
  \end{tabular}
\end{table}

\subsection{Contact Predictor Evaluation}
\label{sec:appendix-contact-pred}

We evaluate the contact predictor described in Section~\ref{sec:appendix-contact-details} on unseen rollouts of the frozen contact-acquisition primitive. We use the 16 training YCB objects under the same domain randomization, with 2560 episodes of 60 steps per hand.

\paragraphc{Evaluation protocol.}
At each control step, \texttt{is\_grasping} is set according to the per-finger contact predictions. We compare this signal against the simulator's ground-truth grasp state. A false positive is a step at which \texttt{is\_grasping} is true while the simulator indicates no grasp, and a false negative is the reverse. At the episode level, a switch is considered correct if the predictor triggers tracking while the hand is actually holding the object.

\paragraphc{Results.}
At a contact-probability threshold of $\tau_c = 0.5$, the per-step false-positive rate is 0.066--0.125 and the false-negative rate is 0.036--0.047 across the three hands. Per-episode switch accuracy is 0.953--0.960, with a median lag of at most one step relative to the simulator's switch. Switch decisions are insensitive to the threshold: switch accuracy remains 0.94--0.98 for $\tau_c \in [0.4, 0.8]$ and degrades only for $\tau_c \le 0.3$.
\begin{figure}[t]
    \centering
    \includegraphics[width=0.5\linewidth]{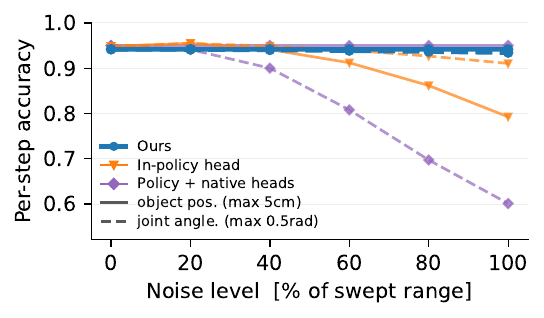}
    \caption{\textbf{Contact-predictor accuracy under observation noise.} Per-step accuracy of the debounced switch signal, averaged over the three hands, as the predictor input is perturbed along one axis at a time (solid: object position, 0--5\,cm; dashed: joint angle, 0--0.5\,rad). Our proprioceptive predictor is unaffected by object-pose noise and degrades least under joint noise.}
    \label{fig:contact-noise}
\end{figure}
\paragraphc{Robustness to observation noise.}
We train a separate proprioception-only predictor because it is more robust to observation noise than a contact head coupled to the policy. To verify this, we compare against two alternatives: a head inside the policy network trained jointly with PPO from the full policy observation, and per-embodiment proprioceptive predictors trained alongside the policy instead of after it. All three are evaluated offline on the same recorded rollouts using their respective inputs and the same ground-truth labels. Noise is added only to the predictor input, one axis at a time (object position up to 5\,cm, joint angles up to 0.5\,rad), while the rollouts themselves are left unchanged. As shown in Figure~\ref{fig:contact-noise}, our predictor is unaffected by object-pose noise by construction and loses at most 0.01 in accuracy under 0.5\,rad joint noise, whereas the in-policy head loses on average 0.16 under object-position noise and the jointly trained heads lose on average 0.35 under joint noise.

\end{document}